%% file: jmlrwcp-sample.tex
\documentclass[pmlr]{jmlr}% new name PMLR (Proceedings of Machine Learning)
\usepackage{booktabs}
\usepackage[load-configurations=version-1]{siunitx}

\theorembodyfont{\upshape}
\theoremheaderfont{\scshape}
\theorempostheader{:}
\theoremsep{\newline}

\usepackage{hyperref}

\let\svthefootnote\thefootnote
\newcommand\freefootnote[1]{%
  \let\thefootnote\relax%
  \footnotetext{#1}%
  \let\thefootnote\svthefootnote%
}
\usepackage{longtable}
\usepackage{afterpage}
\renewcommand{\arraystretch}{1.4}
\usepackage{array}
\newcolumntype{P}[1]{>{\centering\arraybackslash}p{#1}}
\usepackage{changepage}
\usepackage{bbding}
\usepackage{pifont}
\usepackage{stackengine,amssymb,graphicx,scalerel}

\newcommand\BigRec[1]{\stackengine{-.4ex}{\scalebox{.7}{#1}}{\BigCAR}{O}{c}{F}{F}{L}}
\newcommand\BigCAR{\scaleto{\circlearrowright}{3.2ex}}
\newcommand\CAR{\scaleto{\circlearrowright}{2.2ex}}

\jmlryear{2022}
\jmlrworkshop{DYAD'21 Workshop}

\title[Non-verbal social behavior forecasting]{Didn't see that coming: a survey on non-verbal social human behavior forecasting}

   \author{\Name{German Barquero} \Email{gbarquga9@alumnes.ub.edu}\\
   \Name{Johnny Núñez} \Email{jnunezca11@alumnes.ub.edu}\\
   \Name{Sergio Escalera} \Email{sergio@maia.ub.es}\\
     \addr Universitat de Barcelona and Computer Vision Center, Spain
     \AND
   \Name{Zhen Xu} \Email{xuzhen@4paradigm.com}\\
   \Name{Wei-Wei Tu} \Email{tuweiwei@4paradigm.com}\\
     \addr 4Paradigm, Beijing, China
     \AND
   \Name{Isabelle Guyon} \Email{guyon@chalearn.org}\\
     \addr LISN (CNRS/INRIA) Université Paris-Saclay, France, and ChaLearn, USA
     \AND
   \Name{Cristina Palmero} \Email{crpalmec7@alumnes.ub.edu}\\
     \addr Universitat de Barcelona and Computer Vision Center, Spain
     }
\begin{document}

\maketitle

\begin{abstract}
Non-verbal social human behavior forecasting has increasingly attracted the interest of the research community in recent years. Its direct applications to human-robot interaction and socially-aware human motion generation make it a very attractive field. In this survey, we define the behavior forecasting problem for multiple interactive agents in a generic way that aims at unifying the fields of social signals prediction and human motion forecasting, traditionally separated. We hold that both problem formulations refer to the same conceptual problem, and identify many shared fundamental challenges: future stochasticity, context awareness, history exploitation, etc. We also propose a taxonomy that comprises methods published in the last 5 years in a very informative way and describes the current main concerns of the community with regard to this problem. In order to promote further research on this field, we also provide a summarized and friendly overview of audiovisual datasets featuring non-acted social interactions. Finally, we describe the most common metrics used in this task and their particular issues. %We hope this survey can help focus the research efforts on the main challenges of this field.
\end{abstract}
\begin{keywords}
Behavior forecasting, Human motion prediction, Social signal prediction, Social robots, Socially interactive agents, Dyadic interactions, Triadic interactions, Multi-party interactions, Backchanneling, Engagement
\end{keywords}

\input{sections/1_introduction}

\input{sections/2_taxonomy}

\input{sections/3_datasets}

\input{sections/4_metrics}
\input{sections/5_discussion}
\input{sections/6_conclusion}

\acks{
Isabelle Guyon was supported by ANR Chair of Artificial Intelligence HUMANIA ANR-19-CHIA-0022. This work has been partially supported by the Spanish project PID2019-105093GB-I00 and by ICREA under the ICREA Academia programme.
}

\bibliography{jmlr-sample}

\appendix

%\input{sections/7_appendix}
%\section{First Appendix}\label{apd:first}

%This is the first appendix.

%\section{Second Appendix}\label{apd:second}

%This is the second appendix.

\end{document}

%% file: sections/1_introduction.tex
\section{Introduction}

% Paragraph on non-verbal social signals in general.
Communication among humans is extremely complex. It involves an exchange of a continuous stream of social signals among interactants, to which we adapt and respond back. These social signals are manifested as non-verbal behavioral cues like facial expressions, body poses, hands gestures, or vocal feedback.
We, as humans, have the innate capability of identifying, understanding, and processing social cues and signals, which is the core of our social intelligence~\citep{vinciarelli2009social}. Similarly, we are also inherently capable of anticipating, up to some extent, these social signals. For instance, we do not need the speaker to actually end their speech before we know that a turn-taking event is close~\citep{ondavs2019anticipation}. We are prepared in advance. In a similar way, we can anticipate a social action like a handshake by the correct observation and interpretation of simultaneously occurring visual cues from the other interlocutor, like a verbal greeting while their hand is approaching. 
In fact, recent works in neuroscience believe that such anticipation is the motor for cognition. In particular, this current, called \textit{predictive processing}, supports the idea that the brain is constantly generating and updating a mental model of the environment by comparing predicted behaviors to actual observations~\citep{walsh2020evaluating}. Very interestingly, some works have successfully observed interpersonal predictive processing signals during social interactions~\citep{thornton2019social, okruszek2019second}. This suggests the importance that behavior forecasting may have as a pathway to the ultimate behavioral model.
%However, the transfer of such apparently trivial ability to a robotic or virtual agent is extremely difficult.

% Paragraph on importance of behavior forecasting of non-verbal social signals.
If successfully modeled, such forecasting capabilities can enhance human-robot interaction in many applications. For instance, wherever turn-taking events are frequent and very dynamic (e.g., multi-party conversations), any degree of anticipation is extremely beneficial. Being able to anticipate the next speaker or when a listener will disengage are key to efficiently handle such situations. Also, forecasting has direct applications to robot behavior synthesis for social interactions in two directions. First, being able to anticipate the interactants' behavior can help the agent to behave in a more socially-aware way. And second, being able to predict one's own behavior, even by only few milliseconds, can save a valuable computation time. In fact, the longer the future that we are able to predict, the better the robot can prepare for the execution of a movement.

Unfortunately, providing robots or virtual agents with social forecasting capabilities is extremely difficult due to the numerous particularities of the problem. First, the large amount of variables driving a social interaction makes it a highly dimensional problem. For instance, in the previous handshake example, even if the agent detects the approach of the hand as a visual cue and anticipates a handshake, it could miss in the case where the hand is grabbing an object. On top of that, predicting the future always poses problems related to its stochasticity. The plausibility of several equally probable future scenarios complicates the development and evaluation of forecasting models.

In the past years, research on non-verbal social behavior forecasting has followed distinct paths for social signal prediction and computer vision fields, although they share most of their fundamental concerns. For example, the human motion forecasting field does not usually refer to any social signal forecasting work~\citep{mourot2021survey}, even though some of them predict visual social cues, or action labels. And vice versa. This survey wants to unify non-verbal social behavior forecasting for both fields, describe its main challenges, and analyze how they are currently being addressed. To do so, we establish a taxonomy which comprises all methodologies applied to multi-agent (human-human, or human-robot/virtual agent) scenarios and presented in the most recent years (2017-2021). In particular, we focus on works that exploit at least one visual cue. We also engage in a discussion where we foresee some methodological gaps that might become future trends in this field. Besides, we summarize the publicly available datasets of social interactions into a comprehensive and friendly survey. Finally, we present and discuss on the usual evaluation metrics for non-verbal social behavior forecasting.

\begin{figure}[!t]
    \centering
    \includegraphics[width=\textwidth]{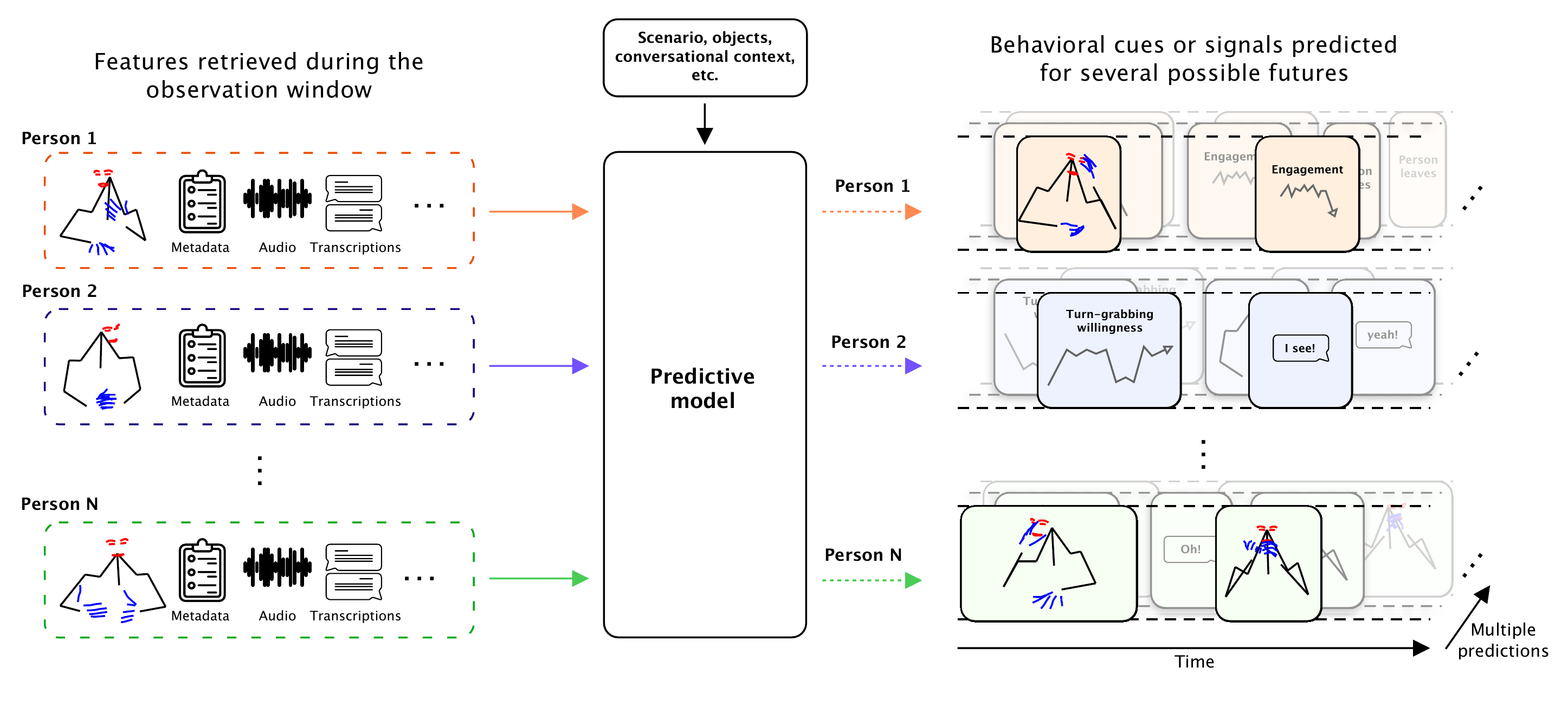}
    \caption{Visual representation of the generic social behavior forecasting problem reviewed in this work. Given a set of features of $N$ multiple interactants observed from the past and the contextual information about the interaction, $M$ future sequences composed of behavioral social cues/signals are predicted.}
    \label{fig:social_signal_forecasting_overview}
\end{figure}

This survey is organized as follows. First, in Section~\ref{sec:taxonomy}, we formulate the non-verbal social human behavior forecasting problem (\autoref{fig:social_signal_forecasting_overview}), and introduce a taxonomy for it. We start by identifying and discussing the main challenges associated to the task, and describe how past works have addressed them (Sections \ref{sec:future} to \ref{sec:framework}). Then, we thoroughly review the state-of-the-art methodologies proposed for non-verbal social behavior forecasting. In particular, we split the survey into methods predicting low-level (e.g., landmarks, facial action units) and high-level (social cues and signals) representations of non-verbal behavior, in Sections~\ref{sec:low_level_representations} and \ref{sec:high_level_representations}, respectively. Then, in Section~\ref{sec:datasets}, an extensive collection of datasets featuring audiovisual social interactions is presented. Datasets are classified according to the interaction scenario (dyadic, triadic, group, or several groups), task (conversational, collaborative, or competitive), and setting (participants are standing or seated). We also provide a summary table that allows the reader to easily compare the low- and high-level behavioral annotations provided as part of each dataset. Section~\ref{sec:metrics} presents the most popular metrics for assessing the accuracy, diversity, and realism of the predicted behavior. In Section~\ref{sec:discussion}, we provide a discussion on general trends and current challenges, as well as possible future research directions for non-verbal social behavior forecasting. Finally, in Section~\ref{sec:ethics}, we review the ethical concerns regarding non-verbal social behavior forecasting and its real-world applications.

%We review:
%\begin{itemize}
%    \item Importance of behavior forecasting for social interactions
%    \item Challenges associated to social interaction forecasting.
%    \item Contributions: 
%    \begin{enumerate}
%        \item survey on social human non-verbal behavior forecasting in interactive scenarios.
%        \item establishment of current state, challenges, and future work of non-verbal social behavior forecasting
%    \end{enumerate}
%\end{itemize}

%% file: sections/2_taxonomy.tex
\section{Taxonomy}
\label{sec:taxonomy}
% Rephrase, we need to make it generalize to low- and high-level representations.

%In order to make an agent more human-like, many works aim at generating human behavior from a single modality such as audio [...]} or Transcripts [...]}. Although such approaches have clear and important applications in embodied conversational agents or virtual assistants, they are using lexico-syntactic information from the future and therefore not predicting the future as such.
%, they are not well suited for latency reduction systems or scenarios where anticipation is critical.

\begin{figure}[!t]
    \includegraphics[width=0.9\textwidth]{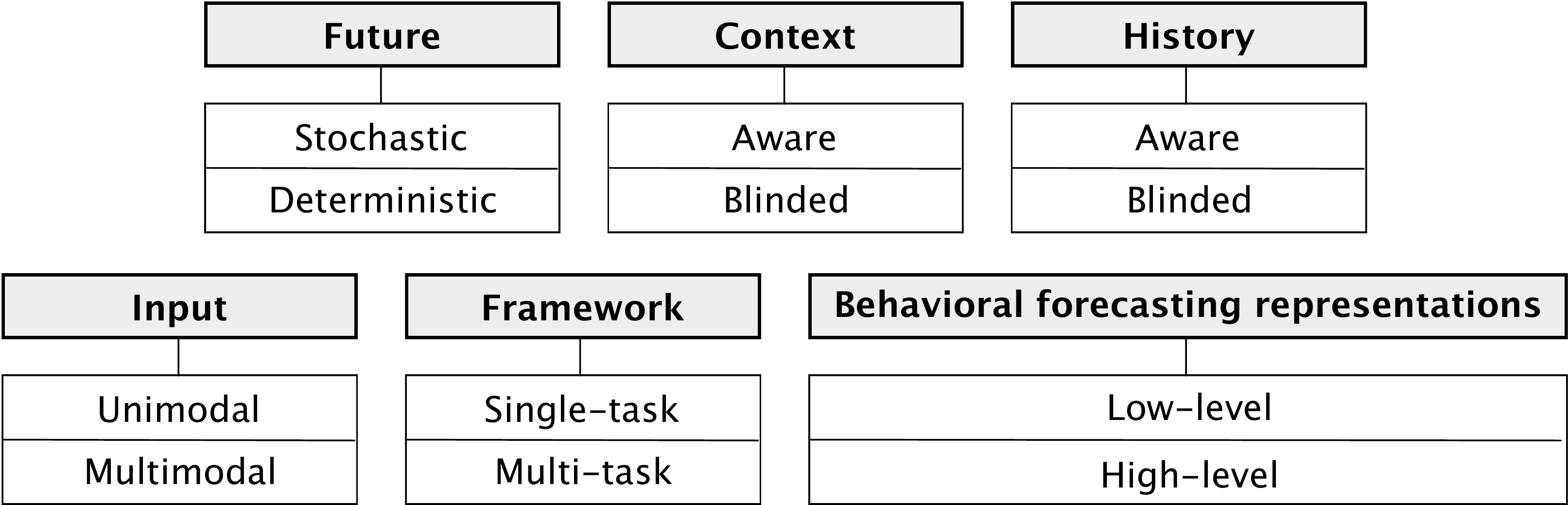}
    \caption{The proposed taxonomy for non-verbal social human behavior forecasting in socially interactive scenarios.}
    \label{fig:taxonomy_scheme}
\end{figure}

Our taxonomy, see \autoref{fig:taxonomy_scheme}, includes all approaches that predict non-verbal human behavior in socially interactive scenarios. These scenarios include at least two subjects socially interacting together, typically referred to as \textit{focused interactions}. % We may need to define what 'socially' means.
Also, we understand forecasting in the strictest sense of the word. Namely, only information observed \textit{before the prediction starts} is used. Such constraints leave co-speech generative methods~\citep{liu2021speech, kucherenko2021large} or pedestrian trajectories forecasting out of our scope. The former leverages the future speech and the latter does not usually feature a focused interaction.
Additionally, the approaches needed to exploit \textit{at least one visual cue} (e.g., landmarks, image, visually annotated behavioral labels). We acknowledge though many works that use exclusively lexical or audio features to predict non-verbal social cues such as backchannel opportunities or turn-taking events~\citep{Ortega2020, Jang2021}. 
On the other hand, the taxonomy is very flexible with regards to the typology of predicted human behavior. Therefore, we include works that predict low-level behavioral representations such as landmarks, head pose or image (\autoref{tab:summary_landmarks}), but also high-level ones like social cues and social signals (\autoref{tab:summary_social_signals})~\citep{vinciarelli2009social}. We encourage the reader to accompany the survey lecture with those tables, as they provide a synthesized view of the methodologies and a comparison among them. In addition, we refer the reader to \autoref{fig:social_signal_forecasting_overview} for an illustration of our definition of behavior forecasting.

Next, we present and describe the main challenges related to non-verbal social behavior forecasting that are currently being actively addressed by the community. They are enclosed within each dimension of our proposed taxonomy: future perspective (Section~\ref{sec:future}), context exploitation (Section~\ref{sec:context}), history awareness (Section~\ref{sec:history}), input modalities (Section~\ref{sec:input}), and framework (Section~\ref{sec:framework}). Finally  (Section~\ref{sec:representations}), we detail and compare related works according to their behavioral forecasting representation (e.g., landmarks, action labels).

\subsection{Future perspective}
\label{sec:future}
A common approach in single-person behavior forecasting consists in embracing the future uncertainty and exploiting it by predicting multiple futures (\textit{multimodal}, or \textit{stochastic})~\citep{aliakbarian2021contextually, hassan2021stochastic, mao2021generating}.
% Comment on how stochastic approaches work in single person motion generation (by citing 1-2 methods) and argue how they would translate to our scenario??
However, this research line has not been fully exploited for social scenarios yet. Instead, most works propose methods which assume that the observed future is unique (\textit{deterministic})~\citep{Adeli2021, Guo2021, Wang2021Multi, Barquero2022}, thus ignoring multiple future behaviors that may co-exist and be equally plausible. This hypothesis removes some challenges associated to stochastic approaches such as the sampling choice among several generated futures, or the assessment of the realism and plausibility of all predicted futures. 
However, this simplification comes at a high cost: the predictive model is penalized for generating plausible and realistic behaviors which do not match the ones observed in the dataset. 
To alleviate this, many works reduce the dimensionality of the forecasting objective (e.g., action labels)~\citep{Sanghvi2019, Airale2021}, or provide extensive contextual information in order to narrow the future space and therefore reduce its stochasticity~\citep{Corona2020, Adeli2020, Adeli2021}. Still, some works forecasting low-level behavioral representations (e.g., landmarks) report a strong regression-to-the-mean effect in the predictions~\citep{Barquero2022}. Some works tried to tackle this problem in several contexts. For example, \cite{Feng2017} proposed building a specific high-frequencies predictor which made the generated facial expressions more realistic. In the context of social signals forecasting, \cite{Raman2021} reasoned that such effect was linked to the availability of similar future signals triggered at different future points. To mitigate it, they proposed to inject the time offset at which social signals triggered into the past encoding. In general though, deterministic works complement the quantitative evaluation with qualitative visualizations that help assess the realism and smoothness of the predictions.

\subsection{Context exploitation}
\label{sec:context}
Social interaction among humans is dynamic and strongly influenced by many external factors. The human behavioral model that drives a conversation between a professor with a student is drastically different from that of a conversation among friends~\citep{reis2000relationship}. Even with the same interactants setting, the place where the interaction happens (e.g., in a bar, at a conference, at home) may change the whole dynamic of their behavior. In a similar way, a handshake might become a handover if the approaching hand is holding an object~\citep{Shu2016,Corona2020}. Although considering all external factors that might influence the behavior is still impractical, some works consider using some contextual information. Accordingly, we split between \textit{context-aware} methods, which were introduced by \cite{Corona2020} and consider at least one modality of contextual information, and \textit{context-blinded} methods, which focus on the target person only~\citep{Wang2021Simple}. Most context-aware methods reviewed in Section~\ref{sec:representations} leveraged the partners' behavioral information. Additionally, other works introduced approaches that also considered the presence and trajectory of objects~\citep{Shu2016, Corona2020, Adeli2021}, 
%the partners' behavior~\citep{Huang2017, Feng2017, Chu2018, Chen2019, Ueno2020, Woo2021,Ahuja2019, Hua2019, Honda2020, Wang2021Multi, Guo2021, Katircioglu2021, Barquero2022}, 
or even the whole visual scene~\citep{Adeli2020}. These methods prove particularly useful in contexts where the behavior is strongly driven by the interaction with the scene.

\subsection{History awareness}
\label{sec:history}
By definition, social interactions evolve over time, generating multiple long-range temporal dependencies. An event at the beginning of an interaction may impact and alter the rest of it. Furthermore, forecasting may benefit from observing very long sequences (e.g., $>$10 seconds) of interactions in order to tune a generic behavioral model to work with the interactants and the specific conversational context.

Although few works attempt to exploit the history in the single-person motion forecasting domain~\citep{Mao2020, Mao2021}, there are few social \textit{history-aware} works. Although they do not detail how long the history can be, \cite{Chu2018} encoded a history of past text sequences and facial expressions with variable length to improve their forecasting capabilities. \cite{Guo2021} and \cite{Katircioglu2021} incorporated motion attention to propagate observed motions to the future, theoretically even when the motion has not been seen in training time. However, both considered small histories of 2 seconds, which only favors the propagation of short repetitive motion. We are not aware of methods that consider much longer historical data, or that learn in an online and adaptive fashion the unique characteristics of each person's behavior.

\subsection{Input modalities}
\label{sec:input}
In addition to the interaction context, the speech, voice tone or other information related to the person of interest or the others may influence their behavior. Naturally, such multimodal data needs to be exploited in a specific way in order to fully profit from it. Therefore, we distinguish between \textit{unimodal} and \textit{multimodal} methods, which combine the visual modality with at least another modality as input to make their predictions.

Most common multimodal settings combined landmarks, body/head pose, or visual cues with past utterance transcriptions~\citep{Chu2018, Hua2019, Ueno2020, Barquero2022}, acoustic features~\citep{Turker2018, Ahuja2019, Ueno2020, Goswami2020, Woo2021, Jain2021, Murray2021, Youssef2021}, speaker's metadata~\citep{Raman2021, Barquero2022}, or with combinations of the previous modalities~\citep{Ishii2020turn, Huang2020, Blache2020, Ishii2021, Boudin2021}. The most common way to exploit different modalities together consists in simply concatenating their embedded representations. Although this has proven to work for several applications, extracting relevant information from multiple modalities is not always straightforward~\citep{Barquero2022}.

\subsection{Framework}
\label{sec:framework}
During the course of an interaction, humans exchange multiple social signals. Turn taking, agreement, politeness, empathy, disengagement, etc, are some examples. In some cases, such signals might be inferred from the same set of social cues, which is the perfect environment for \textit{multi-task} learning. This paradigm, which consists in learning several tasks with the same model, has already helped to improve the results of single task models in many other fields~\citep{zhang2017survey}. In our context, few works have explored multi-task frameworks. \cite{Ishii2020turn, Ishii2021} explored the benefits from predicting several social signals and cues at the same time (turn taking, turn-grabbing willingness, and backchannel responses). \cite{Chu2018} proposed a method to predict the next facial Action Units (AUs) by also predicting the future speech content.

\addtolength{\tabcolsep}{-5pt} 
\begin{table*}[!t]
    \tiny
    \centering
    % the raggedright avoids underfull and auto-justify of multiline text
    \begin{tabular}{>{\raggedright}p{1.4cm}>{\raggedright}p{4.6cm}P{1.5cm}P{1.5cm}P{1.0cm}P{1.0cm}P{1.2cm}P{1.6cm}}
    % the loop symbol was used when they predicted longer time windows by recurrently applying the model. If such recurrency needs 
    % info not available at the beginning of the prediction, they were not considered as such (e.g., 1f).
         \toprule
         Authors & Method highlights & Multimodal input & Context$^\dagger$ & $\rightarrow \mid$ & $\mid \rightarrow$ & Prediction & Scenario \\
         \midrule
         \multicolumn{8}{c}{Face} \\
         \midrule
         % FACE-RELATED PAPERS
         \citealt{Huang2017} & Two GANs. Face image generation from the partner's past expressions. & \ding{55} & Partner & 5s & 1f & FL* & Remote dyadic conv. \\
         \citealt{Feng2017} & VAEs. Parallel low-frequency and high-frequency models favor realism. & \ding{55} & Partner & 3s & \BigRec{0.5s} & FL & Remote dyadic conv. \\
         \citealt{Chu2018} & (Bi-) LSTMs. Multi-task (text+face) setting trained with RL. & Transcripts & Partner & Variable & IPU & AU & Movies dyadic conv. \\
         \citealt{Chen2019} & LSTM+GAN. Face image generation from the partner's past expressions. & \ding{55} & Partner & 0.4s & \BigRec{10} & AU+HP* & TV dyadic conv. \\
         \citealt{Ueno2020} & Bi-GRU. Attention among sequential embeddings of input modalities. & Transcripts +Audio & Partner & Variable & 1f & AU & Triadic conv. \\
         \citealt{Woo2021} & LSTM. Simultaneous prediction for both participants. & Audio & Partner & 0.8s & \BigRec{1} & AU+HP & Dyadic conv.\\
         \midrule
         \multicolumn{8}{c}{Pose (upper or full body)} \\
         \midrule
         % BODY-RELATED PAPERS

         \citealt{Shu2016} & MCMC. Joints functional grouping and sub-events learning. & \ding{55} & Partner +Object & 0.4s & \BigRec{5} & BL & Constrained social actions \\
         \citealt{Ahuja2019} & LSTM/TCN. Dynamically attends to monadic and dyadic behavior models. & Audio & Partner & Variable & 1f & BL & Dyadic conv. \\
         \citealt{Hua2019} & LSTM. Distinct models while speaking (co-speech) and listening (forecasting). & Transcripts & Partner & 2.8s & 2.8s & UBL & Dyadic conv. \\
         \citealt{Honda2020} & LSTM/GRU. Joint encoding and decoding for both interactants. & \ding{55} & Partner & 0.5s & 1s & BL & Competitive fencing \\
         \citealt{Corona2020} & GATs+RNNs. Interactions among objects and subjects modeled. & \ding{55} & Partners +Object & 1s & 2s & BL & Human-object interactions\\
         \citealt{Adeli2020} & GRU. Scene understanding and multi- person encoding (social pooling). & Raw image & Partners +Scene & 0.6s/ 1.3s & 0.6s/ 0.4s & BL & In-the-wild interactions\\
         \citealt{Adeli2021} & GATs+RNNs. Interactions among objects, subjects and scene modeled. & Raw image & Partner +Scene +Objects & 0.6s/1s & 0.6s/1s & BL & In-the-wild interactions \\
         \citealt{Raman2021} & GRU/MLP. Social processes definition and prediction offset injection. & Speaking status & Partners' features & 10f & 10f & HP+Body Pose & Triadic conv. \\
         \citealt{yasar2021scalable} & GRU+Attention. Interpretable latent space. Cross-agent attention. & \ding{55} & Partners & - & 0.6/1.6s & BL & Diverse interactions \\
         \citealt{Wang2021Multi} & Transformer+DCT. Local- and global-range transformers. & \ding{55} & Partners & 1s & 3s & BL & Groups of interactions \\
         \citealt{Guo2021} & Transformer+GCN+DCT. Cross-interaction motion attention (early-fusion). & \ding{55} & Partner & 2s & \BigRec{0.4s} & BL & Dancing interactions \\
         \citealt{Katircioglu2021} & Transformer+GCN+DCT. Pairwise motion attention (late-fusion). & \ding{55} & Partners & 2s & 1s & BL & Dancing interactions \\
         \citealt{Wang2021Simple} & GCN+DCT. Strong and simple baseline with training tricks. & \ding{55} & \ding{55} & 0.6s & 0.6s & BL & In-the-wild interactions \\
         %\midrule
         %\multicolumn{8}{c}{Hands} \\
         %\midrule
         % FULL-BODY RELATED PAPERS
         \midrule
         \multicolumn{8}{c}{Whole body (face+pose+hands)} \\
         \midrule
         \cite{Barquero2022} & LSTM/GRU, TCN, Transformers, and GCN. Weakly supervised with noisy labels. & Audio/ Transcripts /Metadata & Partner & 4s & \BigRec{10}/2s & FL+UBL +HL & Dyadic conv.\\
         \bottomrule
    \end{tabular}
    \caption{\small Summary of papers forecasting low-level representations of non-verbal behavior. All works are history-blinded, with deterministic future, and use at least one visual input modality. Abbreviations: $\rightarrow \mid$,~observation window length; $\mid \rightarrow$,~prediction window length; {\CAR},~Future autoregressively predicted in steps of X frames (or seconds, when specified); 1f, only the immediate next frame is predicted; AU,~action units; HP,~head pose; FL, face landmarks; (U)BL,~(upper) body landmarks; HL,~hands landmarks; LSTM, long short-term memory; Bi, bidirectional; GRU, gated recurrent unit; VAE, variational autoencoder; GAN, generative adversarial network; DCT, discrete cosine transform; TCN, temporal convolutional network; MCMC, Markov chain Monte Carlo; RL, reinforcement learning. *:~Incorporates image generation. $^\dagger$: partners' information used matches the \textit{Prediction} column.}
    \label{tab:summary_landmarks}
\end{table*}
\addtolength{\tabcolsep}{5pt} 

\subsection{Behavioral forecasting representations}
\label{sec:representations}

\subsubsection{Low-level}
\label{sec:low_level_representations}
When interacting with humans, virtual or robotic agents need to be able to reciprocate non-verbal cues at all dimensions. Given the relevance of this problem, many works aim at forecasting low-level representations of non-verbal social cues, see \autoref{tab:summary_landmarks}.
Such representations can be non-semantic representations such as raw image or audio, or semantic ones such as landmarks, head pose, or gaze directions.

% ------------------------ IMAGE AND AUDIO % ------------------------ 
\textbf{Image and audio.} In single human motion forecasting, we find methods that leverage pose motion prediction and generative methods to infer the future image frames~\citep{walker2017pose, zhao2020pose}. There are few works proposing similar two-step approaches for image-based future social behavior forecasting. 
In an interview setting, \cite{Huang2017} proposed a method to generate contextually valid facial images of the interviewer from the past interviewee's facial expressions. To do so, they trained two Generative Adversarial Networks (GAN). The first one, conditioned on the interviewee's recent facial expressions, produced the interviewer expression. The second one was trained to transform the generated expression into a real face image. % Comment limitation of only 1 person considered for the generation => lack of generalization??
\cite{Chen2019} proposed a face-to-face conversation system that also generated real-looking faces in two steps. They designed different models for forecasting behavior during speaking and listening phases. While the former was a co-speech generative method, the latter predicted the future AUs and head pose with a recurrent Long-Short Term Memory (LSTM) unit that only leveraged the past facial gestures of the speaker. Finally, a GAN conditioned on the predicted AUs and head pose generated the face image.
Regarding the prediction of future raw audio output, there are no works that propose such architecture, to the best of our knowledge. A common path is to predict verbal behavior like textual content and include a Text-To-Speech model to generate the speech~\citep{saeki2021incremental}.

% Comment two approaches: window-based and auto-regressive

% ------------------------ FACE % ------------------------ 
\textbf{Face.} Most methods either focus on lower-dimensional representations of the face such as AUs or explicitly learnt representations. Regarding the latter, and aiming at replicating realistic facial gestures, \cite{Feng2017} proposed a Variational Auto-Encoder (VAE) that was trained to explicitly learn a lower-dimensional space for representing facial expressions. This bottleneck helped to reduce the dimensionality of the problem. Then, in order to promote the generation of subtle social cues (e.g., blinking, or eyebrow raising), the encoded past facial expressions of the user and an interactive embodied agent were processed by two specialized predictors, each focusing on either high or low frequencies. Very interestingly, instead of treating it as a regression problem, they clustered the learnt facial latent space and predicted the future expressions in the resulting discrete space. As a result, the regression-to-the-mean effect was mitigated. \cite{Chu2018} also proposed to detach both low- and high-frequency movements generation. Their multimodal model encoded a history representation of past text and facial gestures together (AUs) with the last observed text and facial expression. Then, the future text and the coarse and subtle face expressions were independently predicted in a multi-task setting that was trained with Reinforcement Learning (RL). Finally, they incorporated an adversarial discriminator that promoted the generation of diverse and realistic conversational behavior. \cite{Ueno2020} presented a multimodal approach that embedded text, visual, and audio sequence with bi-directional two-layered GRUs. Then, they were fused by an attention-weighted average layer to predict the face expression of the partner during the immediate feedback response. This visual response was then used to generate the textual feedback, resembling a multi-task setting. Unfortunately, this method cannot be applied to iteratively predict the evolution of the facial expression response in a pure forecasting fashion, as it uses modalities from the immediate previous step as input. Very recently, \cite{Woo2021} described an ongoing research that aims at leveraging audio and the context to predict the future facial expressions and head motion forecasting.

% ------------------------ BODY % ------------------------ 
\textbf{Pose.} The very first attempt to forecast non-verbal body behavior in social interactions was carried out for robot learning of \textit{social affordance}~\citep{Shu2016}. In their work, Shu et al. presented a Markov Chain Monte Carlo (MCMC) based algorithm that iteratively discovered latent subevents, important joints, and their functional grouping. Their method also considered past trajectories of objects to successfully predict the agent's behavior while performing handshakes, high-fives, or object handovers. Favored by the appearance of new and bigger datasets featuring social interactions~\citep{Marcard2018, Andriluka2018, Joo2019}, Recurrent Neural Networks (RNNs) quickly became the standard in human motion forecasting~\citep{Martinez2017, Hua2019, Honda2020}. However, \cite{Honda2020} observed that recurrent models used for single human motion forecasting are not suitable for highly interactive situations like fencing. In their work, they presented a general framework that provided single human motion forecasting methods with the ability to model interpersonal dynamics. To do so, both encoder and decoder LSTMs received as input the previous skeleton (either observed or predicted) concatenated with the hidden state of the opponent in the previous timestep. As a result, the simultaneous behavior forecasting of both players encoded the interpersonal dynamics of the interaction, making the predicted movements more accurate and coherent in the context of competitive fencing. While previous approaches focused on scenarios strongly driven by interpersonal dynamics, \cite{Ahuja2019} emphasized the imbalance between intrapersonal and interpersonal dynamics in dyadic conversations, with considerably less instances from the later. They warned that, in such scenarios, interpersonal dynamics could end up being ignored. To mitigate this issue, they proposed a dyadic residual-attention model (DRAM) that smoothly transitioned between monadic- and dyadic-driven behavior generation. Results showed that their model successfully identified non-verbal social cues like head nod mirroring or torso pose switching and generated proper reactions. \cite{Hua2019} also supported the use of the partner's cues but restricted it to the modeling of the listener's behavior. In their approach, they presented a human-robot body gesture interaction system built similarly to the system of \cite{Chen2019} for facial gestures synthesis. Similarly, they also leveraged two specialized methods for the speaking (co-speech generator) and listening (behavior forecasting) phases of the interaction. In contrast to \cite{Chen2019} though, they incorporated the speaker's speech transcription as an extra predictive feature for the listener's behavior.

\cite{Corona2020} adverted to the fact that, on top of interactions with other humans, human motion is also inherently driven by interactions with objects. To model such interactions, they proposed a method which learnt a semantic graph of human-object interactions during the past observations. Then, the interactions graph was recurrently injected to a RNN in order to generate a context encoding. Both context vector and observed body poses were jointly decoded by a fully connected layer to predict the residuals (motion) of the next pose. As a result, their learnt behavioral model recognized and adapted to the particular dynamics of the scene. Additionally, they proposed to use the context vector to also predict the future motion of the scene objects, and update the context vector accordingly. They reported that their method was state of the art in what scene and human activities understanding refers. With a similar concept in mind, \cite{Adeli2020} proposed an action-agnostic context-aware method. The main difference is that they used spatio-temporal visual features directly extracted from the scene image, so-called context features. Additionally, they introduced a social pooling module that merged the interactants' behavior embeddings in a socially invariant feature vector. Then, the concatenated individual, social, and context features were decoded by a GRU module for each person. Differently from \cite{Corona2020}, the decoding stage did not take into account the interactants' future behavior. In a newer work, \cite{Adeli2021} replaced the social pooling module by a graph attention network (GATs) that modeled interactions among individuals and objects. First, the historic of each person represented as joints-wise attention graph was fed to an RNN to get rid of the temporal dimension. Then, all RNNs outputs were used to build a human-human and a human-object graph attention network, which underwent an iterative message passing algorithm whose flow alternated between both of them. The respective social and context-aware encodings were concatenated to the spatio-temporal visual features and used as the initial hidden state of the RNN-based decoder. In contrast to their prior work, at each person's decoding step, the hidden state was refined by the human-to-human attention graph in order to decode the future motion in a socially aware manner. Similarly to \cite{Corona2020}, they also observed that the socially aware decoding of the predictions improved the overall accuracy.

Very recently, several approaches introduced Transformer-like architectures~\citep{vaswani2017attention} which outperformed previous RNN-based ones~\citep{yasar2021scalable, Wang2021Multi, Guo2021, Katircioglu2021}. \cite{yasar2021scalable} proposed to encode individually the multiple agents' joints positions, velocities, and accelerations. Then, cross-agent attention was applied among the latent space to generate socially aware representations, which followed two subsequent paths. First, these representations went through a two-streams adversarial discriminator that sampled discrete and continuous latent variables. The authors reported that such configuration favored the latent space interpretability. Their analysis on such variables showed that their method effectively captured the underlying dynamics of human motion. Finally, the socially aware latent representations underwent individual recurrent decoders that autoregressively predicted the future sequence of poses. The independent generation of poses represented its main limitation, as generated poses might not be socially coherent.
\cite{Wang2021Multi} proposed to encode local- and global-range dependencies (intra- and inter-personal dependencies, respectively) with two specialized transformer encoders. The past motion of the person of interest was transformed by means of a Discrete Cosine Transform (DCT)~\citep{ahmed1974discrete}, which was then fed to the local-range transformer performing self-attention. At the same time, the global-range transformer encoder applied self-attention across different subjects and different time steps. A spatial positional encoding was added to the global encodings to help the network cluster different individuals in different social interaction groups. Finally, the transformer decoder leveraged the last observed pose as the query, and the local- and global-range encodings as both keys and values in order to generate the whole predicted sequence at once, which was then fed to a linear and an Inverse DCT layer. Additionally, an adversarial loss was used to ensure the realism of the generated behavior. The authors argued that, by predicting the whole motion sequence at once, they prevented generating freezing motion. They reported state-of-the-art and qualitative impressive results in various datasets with several prediction window lengths (up to 3 seconds) and synthetically generated crowded scenarios (up to 15 people).
\cite{Guo2021} provided the motion attention concept originally proposed by \cite{Mao2020} for single human motion prediction with a mechanism to exploit the dyadic dynamics. To do so, they refined the keys and the values of both individuals by applying attention with those of the interactant (cross-interaction attention). The main benefit of motion attention is driven by its capacity of repeating historical patterns even for longer observed windows than the ones used for training. In a highly interactive scenario like dancing, they showed quantitative and qualitative improvements over the naive adaptation of their base method to interactive scenarios (\citealt{Mao2020}'s method with concatenation of inputs).
Very similarly, \cite{Katircioglu2021} recently presented an analogous adaptation of \cite{Mao2020}. Instead of refining each interactants' keys and values with the others', \cite{Katircioglu2021} suggested having two branches to exploit the single and multi-person dynamics through self-attention and pairwise attention, respectively, and merge them after the decoding stage. Leveraging the interactant's motion relative to the person of interest's coordinates helped to model the interaction. Similarly to the cross-interaction attention, the pairwise attention also outperformed the concatenation-based base method and provided much more interactive predictions. As its main limitation, they raised the point that the fact that each subject has their own dancing style might sometimes cause unsatisfactory results.
Curiously, the three state-of-the-art transformer-based methods integrated the DCT to predict a whole motion sequence at once in a non-recurrent manner to avoid freezing motions. This already devises a future trend in low-level behavioral representations forecasting.
% to jointly predict human pose and the trajectory

In contrast to the previous highly complex approaches, \cite{Wang2021Simple} recently proposed an unimodal and context-blinded method which beat its multimodal and context-aware competitors in a multi-person motion prediction benchmark~\citep{Adeli2020, Adeli2021}. They used the work of \cite{Mao2019} as backbone, which consisted of cascaded Graph Convolutional Networks (GCNs) applied to the DCT of the joints. They proved that using several training tricks such as interpolation of invisible/missing joints, data augmentation, boundary filtering, or curriculum learning, among many others, may be more effective than leveraging more complex networks.

% ------------------------ HANDS % ------------------------ 
\textbf{Hands.} Even though anticipating the hands' motion and gestures might be useful for social behavior modeling, we did not find any work within the scope of our survey. Most related work on hands focus on human-object affordance~\citep{lee2018forecasting, corona2020ganhand}, or hands motion prediction in non-social contexts~\citep{luo2019human}.

% ------------------------ FULL BODY % ------------------------ 
\textbf{Whole body.} Few works have attempted to jointly model the behavior of body and face~\citep{Grafsgaard2018, Joo2019}. However, they do not fall within the scope of this work as all of them used future information of either another modality (e.g., text, speech) or the interactant. Very recently, a behavior forecasting competition leveraging whole-body landmarks was held within the ChaLearn LAP DYAD@ICCV'21 workshop~\citep{Palmero2022}. The common trend observed during the competition coincides with the classic path for body pose forecasting: recurrent encoder-decoder architectures with adversarial losses that ensure realism. Although none of the teams beat the competition baseline, the organizers identified some of the main challenges. The usage of noisy labels, the highly stochastic nature of the hands, or the mostly static nature of the dataset (seated dyadic conversations) are some examples. Motivated by this workshop's benchmark, \cite{Barquero2022} proposed several state-of-the-art methodologies that outperformed the competition's baseline. Consistently to the recent findings in body pose forecasting, they also found that Transformer-like architectures provided the best results in whole-body behavior forecasting. Interestingly, their best results were obtained by only leveraging the prediction of one part of the body at a time (face, pose, and hands). They hypothesized that it could be due to the significant behavioral differences among parts of the body. They also underlined the need of larger datasets to model such high dimensional problems.

\subsubsection{High-level}
\label{sec:high_level_representations}

The ability to understand social signals or behaviors lies in the correct detection of their several distinctive associated social cues~\citep{vinciarelli2009social}. Therefore, their \textit{early-detection} or \textit{anticipation} is of utmost importance in many social applications~\citep{ondavs2019anticipation}. Both social cues and signals are comprised within our definition of high-level representations of non-verbal behavior, see \autoref{tab:summary_social_signals}. Note that our survey includes works aiming at predicting such behavioral representations at \textit{any} time in the future. This also includes works aiming at the \textit{immediate future} (e.g., decision making, behavior generation). This distinction is noticeable by looking at the future length column in \autoref{tab:summary_social_signals}.

\addtolength{\tabcolsep}{-5pt} 
\begin{table*}[t!]
    \tiny
    \centering
    % the raggedright avoids underfull and auto-justify of multiline text
    \begin{tabular}{>{\raggedright}p{1.7cm}>{\raggedright}p{4.5cm}P{1.6cm}P{1.2cm}P{1.0cm}P{1.0cm}P{2.0cm}P{1.5cm}}
    % info not available at the beginning of the prediction, they were not considered as such (e.g., 1f).
         \toprule
         Authors & Method highlights & Multimodal input & Context & $\rightarrow \mid$ & $\mid \rightarrow$ & Prediction & Scenario \\
         \midrule
         \multicolumn{8}{c}{Social cues} \\
         \midrule
         %\midrule
         %\multicolumn{6}{l}{Vocal social cues (audible)}} \\
         %\midrule
         %\citealt{Boudin2021} & Logistic regression. Multimodal prediction of feedback subtypes. & Visual +Acoustic +Text (word) & Listener's features & IPU & Imm. & Backchannel opportunity +subtypes & Dyadic conv. \\
         %\citealt{Ishii2021} & MLP. Multimodal multi-task framework for backchannel opportunity prediction. & Visual +Acoustic +Text (utterance) & Listener's features & IPU & Imm. & Backchannel opportunity & Dyadic conv. \\
         %\citealt{Jain2021} & LSTM. Semi-supervised annotation method. & Visual +Acoustic & Listener's features & 3s & 3s & Verbal/visual backchannel opportunity & Dyadic conv. \\
         %\citealt{Jang2021} & Bi-LSTM. Multi-tasking: sentiment scores and BCs' category prediction. & Acoustic+ utterances & Partner's utterance +speech & 1.5s & Imm. & Backchannels & Psychological counseling \\
         %\citealt{Adiba2021} &  & Utterance as words &  & Whole & 0-2s & Backchannels \\
         %\citealt{Blache2020} & Rule-based. It leverages multimodal features at a discourse level. & Visual +Acoustic +Text (discourse) & Listener's features & Discourse length & Imm. & Verbal/visual backchannel opportunity & Doctor - patient dialogs \\
         %\citealt{Ortega2020} & CNN. Personalised listener's BC generation. & Acoustic+ words & Partner's utterance +speech & 2s & Imm. & Backchannels \\
         %\citealt{Park2017} &  & Non-verbal behavior &  & 2.7s & Imm. & Backchannels \\

         %\midrule
         %\multicolumn{6}{l}{Visual social cues (observable)}} \\
         %\midrule
         \citealt{Sanghvi2019} & GRU+Attention. Social signals are gated with social attention. & \ding{55} & Partners' features & 15 actions & Imm. & Behavioral actions & Simulated group interactions \\
         \citealt{Huang2020} & SVM. Multimodal generation of nodding and attention behavior from listeners' behavior only. & Acoustic +Transcripts (words) & Listeners' features & 1-10s & Imm & Nodding + attention & Multi-party meeting (HRI) \\
         \citealt{Blache2020} & Rule-based. It leverages multimodal features at a discourse level. & Acoustic +Transcripts (discourse) & Listener's features & Discourse length & Imm. & Backchannel opportunity +subtypes & Doctor - patient dialogs \\
         \citealt{Jain2021} & LSTM. Semi-supervised annotation method. & Acoustic & Listener's features & 3s & 3s & Verbal/visual backchannel opportunity & Dyadic conv. \\
         \citealt{Ishii2021} & MLP. Multimodal multi-task framework for backchannel opportunity prediction. & Acoustic +Transcripts (utterance) & Listener's features & IPU & Imm. & Backchannel opportunity & Dyadic conv. \\
         \citealt{Boudin2021} & Logistic regression. Multimodal prediction of feedback subtypes. & Acoustic +Transcripts (word) & Listener's features & IPU & Imm. & Backchannel opportunity +subtypes & Dyadic conv. \\
         \citealt{Murray2021} & LSTM. Useful data augmentation techniques. & Acoustic & Partner's features & 2s & Imm. & Nodding & Remote dyadic conv. \\
         \citealt{Airale2021} & LSTM+GAN. Pooling module and dual-stream discriminator. & \ding{55} & Partners' actions & 3s & 3s &  Behavioral actions & Cocktail party \\
         \citealt{Raman2021} & GRU/MLP. Social processes definition and prediction offset injection. & Speaking status & Partners' features & 10f & 10f & Speaking status & Triadic conv. \\
         %\citealt{Raman2021} & GRU/MLP. Social processes definition and prediction offset injection. & Speaking status & Partners' features & 10f & 10f & Head/body pose /Speaking status & Triadic conv. \\
         %\citealt{Ozkan2013} &  & Prosody +Verbal &  & ?? & 1s & Nodding \\
         %\citealt{Morency2010} &  & Speech +Gestures &  & 2s & Imm & Nodding \\

         \midrule
         \multicolumn{8}{c}{Social signals} \\
         \midrule
         \citealt{Ishii2017} & SVM. Speaker's and listeners' head motion and synchrony leveraged. & \ding{55} & Listeners' head pose & 3s & 0-4s & Next-utterance timing & Multi-party meeting \\
         \citealt{Turker2018} & SVM+LSTM. Fusion of classifiers. Many prediction window lengths tested. & Acoustic & Partner's features & 3s/2s & 0-1s & Nodding/Turn-taking & Dyadic conv. \\
         \citealt{VanDoom2018} & AdaBoost. Forecasting when an engagement breakdown will occur. & \ding{55} & Partners' features & 10s & 0-30s & Engagement breakdown & Cocktail party \\
         \citealt{Ishii2019} & SVM+SVR. Three-step model that combines mouth and gaze visual cues. & \ding{55} & Listeners' mouth + gaze cues & 1.2s & 2s & Next speaker + utterance interval & Multi-party meeting \\
         \citealt{Ishii2020turn} & MLP. Multimodal multi-task framework for turn-changing anticipation. & Acoustic +Transcripts (utterance) & Listener's features & IPU & +/- 0.6s & Turn-taking & Dyadic conv. \\
         \citealt{Goswami2020} & Random Forests / ResNet. Visual cues leveraged for the first time for this task. & Acoustic & Speaker's acoustic features & 3s & 3s & Disengagement & Storytelling to children \\
         \citealt{Muller2020} & LSTM. Multimodal encoding predicts gaze aversion probability. & Acoustic & Partner's features & 6.4s & 0.2-5s & Gaze aversion & Remote dyadic conv.\\
         \citealt{Youssef2021} & Logistic regression. Combinations of multiple modalities explored. & Acoustic & Partner's features & 10s & 5s & Engagement breakdown & Dyadic interaction (HRI) \\
         
         %\citealt{Bohus2014} &  & Visual +speech &  & 0.2-8s &  & Engagement breakdown \\
         \bottomrule
    \end{tabular}
    \caption{Summary of papers forecasting high-level representations of non-verbal behavior. Abbreviations: IPU, inter-pausal unit; Imm., immediate future; conv., conversation; HRI, human-robot interaction; GRU, gated recurrent unit; MLP, multi-layer perceptron; LSTM, long short-term memory; SVM: support vector machine; GAN, generative adversarial network; SVR, support vector regression.}
    \label{tab:summary_social_signals}
\end{table*}
\addtolength{\tabcolsep}{5pt}

\textbf{Social cues}. %Acoustic features prediction? This can be speech pause, long pause, prosody changes, etc. 'Mmmh', "Well..." 
Backchannel responses are among the most explored social cues\footnote{The definitions of social cues and signals used in this work are borrowed from the Social Signal Processing domain~\citep{vinciarelli2009social, Raman2021}. Accordingly, a \textit{social signal} refers to the relational attitudes displayed by people. Such signals are high-level constructs resulting from the cues perception.} in human-robot interaction scenarios. Such cues can be vocal (e.g., 'Mmmh!', 'Well...!') or visual (e.g., head nodding), and are of utmost importance in order to keep the interacting user engaged~\citep{krauss1977role}. Earlier classic approaches built handcrafted sets of rules that triggered generic backchannel responses~\citep{al2009generating, poppe2010backchannel}. %\cite{Ortega2020} presented a multimodal architecture that combined embeddings of the listener identifier and the speaker's acoustic (Mel-frequency cepstral coefficients) and lexical word embeddings in order to predict the BC category (continuer, asessment, and No-BC). Although the listener's BC personalisation is novel and effective, it does not suit a real deployment where the model deals with unseen listeners.
%In an iterative improvement to the work of \cite{Ortega2020}, \cite{Jang2021} proposed to extract lexical information at the speaker's utterance level instead. With this small paradigm shift, they already improved the overall accuracy by 11.6\%. Such improvement is especially notable for empathic response (17.1\%), which might be driven by the superior semantic content of the utterance embedding compared to the word embedding's. Additionally, they proposed a novel multi-task setting in which they also predicted the sentiment scores associated to the speaker's utterances. The sentiment cues recognition helped to increase the BC prediction performance an additional 2.7\%.
In fact, forecasting backchannel subtypes (generic, agreement, disagreement, surprise, fear, etc.) had traditionally required different levels of semantic processing. \cite{Blache2020} proposed a novel single-route backchannel predictive model that revisited the rule-based classic paradigm and predicted backchannels in real time at a fine-grained level. Their method used prosodic, discourse, semantic, syntax, and gesture features. In contrast with previous approaches that used an observation window as long as the last utterance, they proposed to extract features from bigger semantic units by means of discourse markers.
More recently, the collection, annotation and release of bigger datasets favored the appearance of data-driven automated multimodal methods for backchannel prediction. 
For example, \cite{Boudin2021} used a logistic classifier that was trained on %for the first time 
visual cues, prosodic and lexico-syntactic features in order to predict not only the backchannel opportunity but also their subtype associated (generic, positive, or expected). The choice of such a simple classifier was driven by the small dataset available. They showed the superior performance of the multimodal combinations in both tasks.

Other works also focused on the visual dimension of backchannel responses. For example, in a human-robot interaction, \cite{Huang2020} proposed a multimodal Support Vector Machine (SVM) that fused prosodic, verbal (word-based), and visual (head motion and gaze attention) features from only the human interlocutors to generate behavior based on nods and gaze attention switches. They argued that different behaviors needed to be modeled for the three possible situations: speaking, listening, and idling (while no one speaks). Even though their results show a fair prediction capability, the model was only tested for an immediate predicted reaction (next frame). We expect the model to struggle with an earlier anticipation as no intentional or behavioral information was encoded. In fact, the authors claimed that the generation of fully autonomous behavior in groupal human-robot interactions is still beyond their capabilities.
One of the main reasons behind such pessimistic point of view lays on the numerous particularities of behavior forecasting. For example, backchanneling periods in a conversation are short and infrequent, which leads to a huge imbalanced problem. %As a result,  when predicting backchannels is data scarcity, as the . 
\cite{Murray2021} proposed a data augmentation method that tried to mitigate this issue when predicting head nodding. The data augmentation focused on the frequency acoustic features and consisted in warping them over time, masking blocks of utterances, and masking blocks of consecutive frequency channels. A similar technique was explored for the head pose, which was also warped in space and time, and masked over time. The experiments showed important improvements when forecasting head nodding with the combination of these strategies and an LSTM. Another big challenge is the extremely time-consuming annotation of social datasets. In the backchannel scenario, the highly multimodal nature of an interaction requires the annotator to pay attention to the audio, speech, and visual content before making a decision. This process is tedious and prone to errors.
Very recently, \cite{Jain2021} proposed a semi-supervised method for identification of listener backchannels that was able to detect up to the 90\% of the backchannels with only a small subset of labeled data (25\%). More importantly, it identified the type of the signals associated around 85\% of the times. The authors showed that models trained on such noisy labels were able to keep a 93\% and a 96\% of the performance with respect to those trained with the cleaned annotations for the tasks of backchannel opportunity prediction and signal classification, respectively. Their general methodology can be adapted for other conversational datasets. Although its validation in other datasets is still pending, it represents an important first step to speed-up the annotation processes and reduce the workforce that they require.

Clearly, the prediction of backchannel responses has attracted a lot of attention. However, they are not the only type of non-verbal behavioral social cues. In a more general framework, few works have tried to predict the future development of an interaction leveraging low-level action labels (e.g., speaking, idling, laughing), motivated by the recent release of annotated audiovisual datasets~\citep{AlamedaPineda2016SALSAAN, CabreraQuiros2018TheMD}.
\cite{Sanghvi2019} proposed a model that used the visual features (e.g., location, gaze orientation) from all interactants to predict the target's future actions (e.g., speak, listen, leave). To do so, a GRU encoded the features of each interactant concatenated with those from the target person. Then, similarly to the attention-based methods reviewed in Section~\ref{sec:low_level_representations}, their method applied attention across all social encodings, which were then fed to a pooling layer so that an arbitrary number of individuals could be handled. Finally, two dense layers converted the output of the pooling layer to the probability distribution defining the next conversational action. They reported that, thanks to the social attention, group annotations were not required. % They concluded that the collection of larger-scale datasets will eventually help to predict more complex behaviors.
\cite{Airale2021} also defined the non-verbal behavior forecasting as a discrete multi-sequence generation problem. Their methodology was radically different though: a GAN conditioned on the observed interaction, which was encoded with an LSTM. In the generative stage, a socially aware hidden state was computed at each timestep. The strategy consisted in using a pooling module to update the hidden states of each person's LSTM decoder and convert them to new socially aware LSTM hidden states for the next decoding step. As a result, the decoded actions were obtained in a coherent way with respect to the actions generated for all surrounding subjects. Additionally, they presented a two-streams novel adversarial discriminator. The first branch corresponded to the classic one, so it favored the realism of individual action sequences disregarding any contextual information. The second one combined the predicted actions sequences with a pooling module of all individuals in the scene to ensure that the generated interactions as a whole were realistic. As a result, the consistency across the predicted action sequences for all participants of the interaction was preserved.

\textbf{Social signals.} The ability to recognize human social signals and behaviors like turn-taking, disengagement, or agreement, and act accordingly are the keys to develop socially intelligent agents~\citep{vinciarelli2009social}. There are many other human capabilities which are carried out unconsciously, like anticipation. For example, a person starts building their response to the speaker's speech before their turn is over, thanks to anticipating the turn taking event~\citep{ondavs2019anticipation}. Consequently, research has focused on trying to find the most important social cues when it comes to anticipating the appearance of social signals of interest. For example, \cite{Ishii2017} found that the amount of head movement of the current speaker, next speaker, and listeners had fairly good prediction capabilities with regards to the next utterance timing. They proposed a light SVM method that could be deployed to any agent equipped with a camera or a depth sensor. Similarly, in multi-party conversations, \cite{Ishii2019} discovered that the speaker's and listeners' mouth-opening transition patterns could be used to predict the next speaker and the time interval between the current utterance ends and the next utterance begins. To prove it, they developed a three-step system. First, an SVM model predicted whether a turn-changing would be produced next. If the answer was yes, then another SVM predicted the next speaker. Finally, independent SVR models for turn-changing and turn-keeping events predicted the utterance interval. The good results of the predictive model suggested the importance of visual cues in forecasting the conversational flow. Its exploitation could potentially help conversational agents to raise the participants' engagement before the start of the next utterance. 

Actually, turn-taking modeling has always attracted a lot of attention. An appropriate turn management is very much needed for a smooth and fluent conversation, which is determining for a pleasant human-robot interaction.
%\cite{Ishii2019} discovered different mouth-opening transition patterns for the current speaker and listeners depending on the next speaker and the time interval between current utterance ends and next utterance begins in multi-party conversations. To prove it, they developed a three-step system that combines the speaker's and listeners' gaze targets with mouth-opening labels. First, an SVM model predicted whether a turn-changing will be produced next. If the answer is yes, then another SVM predicts the next speaker. Finally, independent SVR models for turn-changing and turn-keeping events predict the utterance interval. The good results of the predictive model suggest the importance of visual cues in forecasting the conversational flow and will help conversation agents to engage participants before the start of the next utterance.
In this line, \cite{Turker2018} made one of the first attempts to exploit multiple modalities (acoustic features and visual cues) to predict turn-taking. Among their contributions, they presented an approach that summarized each acoustic feature into a set of statistical measures across the temporal axis (e.g., mean, deviation, skewness). As a result, thanks to the removal of the temporal axis, a simple SVM could be used as classifier. Unfortunately, their tests with turn-taking and head-nodding behavior forecasting showed a superior performance of the recurrent LSTM alternative, thus proving the relevance of the temporal dependencies for such tasks. They also showed that, while the unimodal (acoustic features) forecasting results were close to random, the multimodal performance was promising. 
Many subsequent works have presented successful multimodal methodologies for forecasting other social signals. For example, \cite{Ishii2020turn} analyzed the relationship between turn-holding and grabbing willingness and the actual turn-change. Although they found discrepancies between willingness and actual turn-changing behavior, building a multimodal multi-task model to simultaneously predict both turn-willingness and turn-changing behaviors improved the results for both tasks. In a posterior work, \cite{Ishii2021} expanded their multimodal multi-task framework with the backchannel opportunity prediction task. They showed that, while backchannel prediction benefited from the multi-task learning, no improvement came from adding the turn-changing prediction. Among their conclusions, they stated that, in both cases, simultaneously employing features from the speaker and the listener helped to improve the predictions. One of their main limitations though was the limited exploration of fusion strategies for the modalities used, as their choice was simply concatenating the three feature vectors of audio, text, and video, before a dense layer.

% Disengagement
Disengagement is another very important social signal to be considered when designing interactive agents. If identified with enough anticipation, the speaker can prevent it from happening with backchannel responses or by making an engaging hands gesture, for example.
\Citet{VanDoom2018} made an attempt to anticipate whether a user would leave the conversation and when. Unfortunately, in the first task, they only slightly outperformed the random baseline with one of the many models trained (AdaBoost), and were not successful at the second.
Similarly in a human-robot interaction scenario, \cite{Youssef2021} aimed at anticipating a premature ending of an interaction. As part of their experiments, a logistic regression classifier was trained with all the possible multimodal combinations. They found that the best results were achieved with the combination of the distance to the robot, the gaze and the head motion, as well as facial expressions and acoustic features. Surprisingly, the choice of the classifier had little effect over the final results (logistic regression, random forest, multi-layer perceptron, or linear-discriminant analysis). The small influence of the classifier choice has been consistent along all works reviewed in this section. We hypothesize that this is due to the little need of further processing for such simple and already high-level representations.
As a matter of fact, \cite{Goswami2020} found few performance differences between a random forest and a ResNet when predicting disengagement in the context of storytelling with children, who are prone to disengage very easily. In this work, they also predicted whether a low or a high degree of backchanneling was needed to keep the listener engaged. They assessed the prediction capabilities of many visual cues never used before for this task such as pupil dilation, blink rate, head movements, or facial AUs. Interestingly, they found that the gaze features and the speech pitch were among the most important features to be considered for disengagement prediction. 
These findings are consistent with those from the work of \cite{Muller2020}, who found eye contact and speaker turns to be the most informative modalities when it comes to anticipating averted gaze during dyadic interactions. In their work, the authors also tested other less powerful modalities including face- and gaze-related attributes, expressions and speaker information.

As observed in the course of the survey, there is an important heterogeneity with respect to methodologies used for social signal/cues forecasting. \cite{Joo2019} tried to establish a generic definition which homogenized all methodologies: a \textit{social signal prediction} (SSP) model. The SSP model defined a framework to model the dynamics of social signals exchanged among interacting individuals in a data-driven way. It consisted in using the target's past behavior and their interactants' to predict its future behavior. However, their definition implied that a separate function was learnt for every person. \cite{Raman2021} solved this issue in their formulation of \textit{social processes} (SP), in which a simultaneous prediction of the behavior of all the individuals was considered.

%WTF?? As a generalization for social signal processing in the end??} Sometimes, a single observation might have multiple associated future signals at different offsets. \cite{Raman2021} report that this is one of the reasons behind the commonly observed averaging effect. To mitigate it, they proposed to inject the offset into the encoded vector. %Additionally, any error propagation derived from making recurrent predictions until the offset is covered is also avoided with this method.}

%Finish with small paragraph summarizing high-level forecasting and comparing to low-level. We may even relate with datasets???}

%which presented a multimodal framework to anticipate averted gaze during dyadic interactions. To do so, they encoded face- and gaze-related attributes, expressions and speaker information to output a single prediction for a specific future window length. Surprisingly, their ablation on the combination of inputs shows that the method leveraging only eye contact and speaker turns has predictability capabilities comparable to those of the full method.

%% file: sections/3_datasets.tex
\section{Datasets}
\label{sec:datasets}

\begin{figure}[!t]
    \centering
    \includegraphics[width=0.8\textwidth]{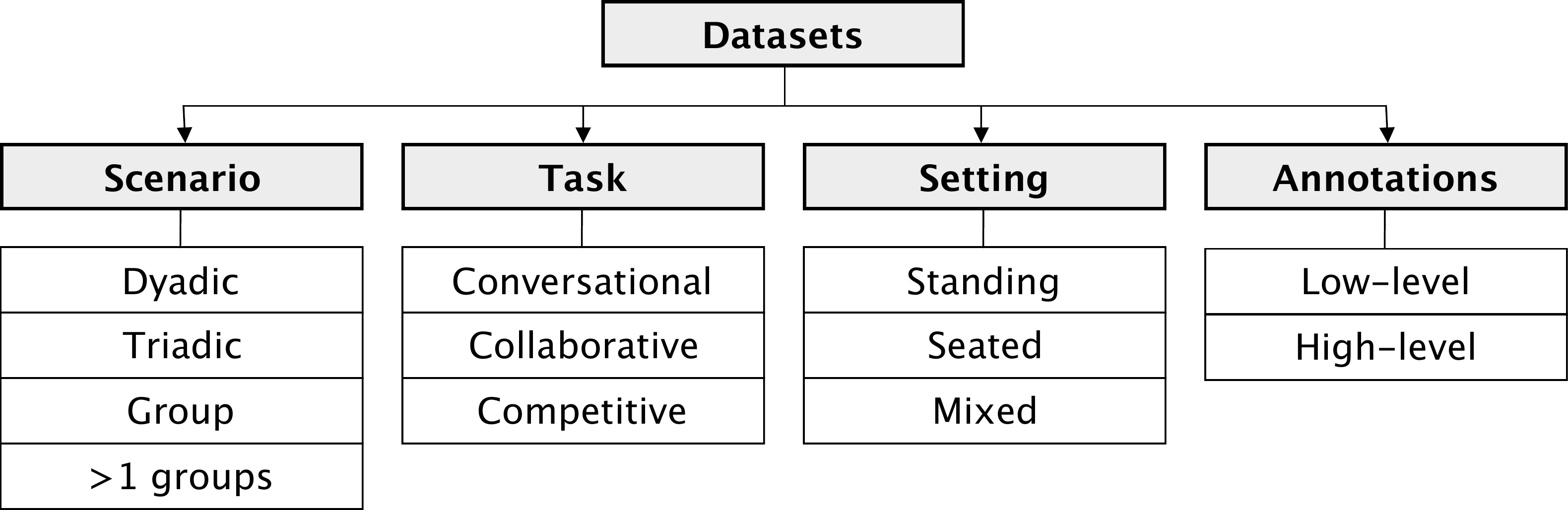}
    \caption{Classification of audiovisual datasets featuring non-acted social interactions.}
    \label{fig:datasets_taxonomy}
\end{figure}

\begin{figure}[!t]
    \centering
    \includegraphics[width=\textwidth]{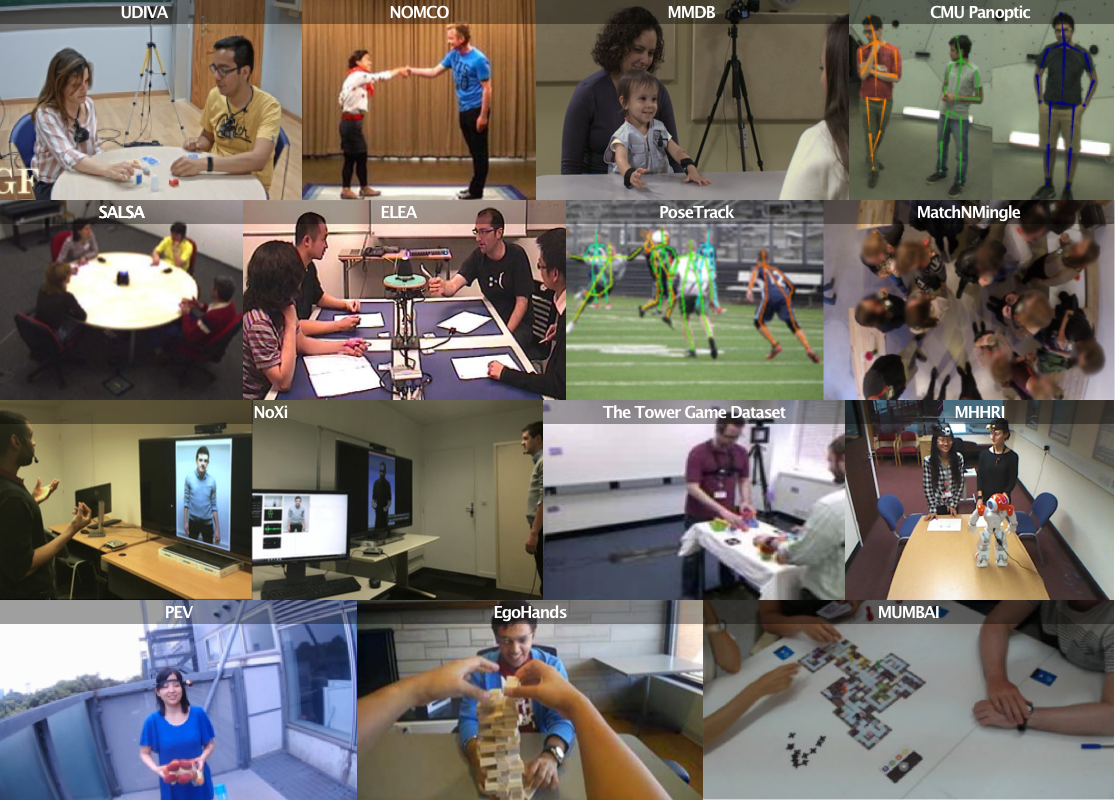}
    \caption{Samples representative of the types of dataset scenarios included in this survey.}
    \label{fig:datasets_image}
\end{figure}

A recurrent problem observed in our survey and one of the main challenges of non-verbal social behavior forecasting is the lack of large annotated datasets. In order to provide the reader with an overview of the currently available datasets, we briefly go through them in this section. Note that we restrict the survey to \textit{publicly available} datasets that feature \textit{audiovisual non-acted social interactions}. The taxonomy that we present, see~\autoref{fig:datasets_taxonomy}, groups them into scenarios (dyadic, triadic, group, and $>$1 groups), tasks (conversational, collaborative, and competitive), and settings (standing, or seating) that elicit different behavioral patterns. \autoref{fig:datasets_image} shows illustrative examples of the scenarios considered in this part of the survey.
%In order to cluster them into interactions favoring the appearance of similar behavioral patterns, we established a datasets taxonomy presented in \autoref{fig:datasets_taxonomy}.

\renewcommand{\arraystretch}{1.3}
\addtolength{\tabcolsep}{-5pt} 
\begin{table}[!t]
\tiny
    \centering
    \begin{tabular}{>{\raggedright}p{1.7cm}P{1.5cm}P{1.0cm}P{1.0cm}P{1.0cm}P{1.0cm}P{1.0cm}P{1.0cm}P{2.2cm}P{3.0cm}}
         \toprule
         & Dataset & Scenario & Task & Setting & Content & \#Subjects & Size & \multicolumn{2}{c}{Behavioral Annotations} \\
          &  &  &  &  & & & & Low-level & High-level \\
         \midrule
         \multicolumn{10}{c}{Third-person view} \\
         \midrule
         % TODO
         \citealt{McCowan2005TheAM} & AMI & Group & Conv. & Seated & A,T & ? & 100h & \ding{55} & Turn Taking, Gestures, Emotions, Game Decisions \\
         \citealt{DouglasCowie2007TheHD} & HUMAINE & Multiple & Multiple & Seated & A,P,T & 309 & $>$26h & Face Expression & Face Gestures, Emotions \\
         \Citealt{vanSon2008TheIC} & IFADV & Dyadic & Conv. & Seated & A & 34 & 5h & Gaze & Emotions, Turn talking, Feedback Responses \\
         \citealt{Bertrand2008LeC} & CID & Dyadic & Conv. & Seated & A & 16 & 8h & \ding{55} & Phonetics, Prosody, Morphology, Syntax, Discourse, Face Gestures \\
         \citealt{Edlund2010SpontalAS} & Spontal & Dyadic & Conv. & Seated & A & ? & 60h & Motion Capture & \ding{55} \\
         \citealt{Hung2010TheIW} & IDIAP Wolf & Group  & Comp. & Seated & A & 36 & 7h & \ding{55} & Speaking Segments, Deceptive/Non-Deceptive Roles, Game Decisions \\
         \citealt{Lcking2012DatabasedAO} & SaGA & Dyadic & Conv. & Seated & A,T & 50 & 4.7h & \ding{55} & Gestures \\
         \citealt{Soomro2012UCF101AD} & UCF101 & Multiple & Multiple & Mixed & A & ? & 27h & \ding{55} & Action labels \\
         \citealt{SanchezCortes2012ANB} & ELEA & Triadic, Group & Collab. & Seated & A,Q & 102 & 10h & \ding{55} & Power, Dominance, Leadership, Perceived Leadership, Competence, Likeness \\
         \citealt{Rehg2013DecodingCS} & MMDB & Dyadic & Conv.+ Collab. & Seated & A,P & 121 & $\sim$10h &  Face Expressions, Gaze & Vocalizations, Verbalizations, Vocal Affect, Gestures \\
         \citealt{Vella2013OverlapsIM} & MAMCO & Dyadic & Conv.+ Collab. & Standing & A,P,T & 12 & $\sim$1h &  \ding{55} & Turn Overlap \\
         \citealt{Bilakhia2015TheMM} & MAHNOB & Dyadic & Conv. & Seated & A & 60 & 11.6h &  Face Expressions, Head, Body, and Hands Motion, Postural Shifts  & \ding{55}\\
         \citealt{Vandeventer20154DCC} & 4D CCDb & Dyadic & Conv. & Seated & A,T & 4 & $\sim$0.5h &  Face Expressions, Head Motion, Gaze & Back/Front Channelling, (dis)Agreement, Happiness, Surprise, Thinking, Confusion, Head Nodding/Shaking
         /Tilting \\
         \citealt{Salter2015TheTG} & The Tower Game & Dyadic & Comp. & Standing & A & 39 & 9.5h & Face Landmark, Gaze, Person Tracking & \ding{55} \\
         \citealt{naim2015automated} & MIT Interview & Dyadic & Conv. & Seated & A,T & 69 & 10.5h & Face Expressions & Friendliness, Presence, Engagement, Excitement, Focused, Calm, Authentic \\
         \citealt{Shukla2016MuDERIMD} & MuDERI & Dyadic & Conv. & Seated & A,B,P & 12 & $\sim$7h & \ding{55} & Valence, Arousal \\
         \citealt{AlamedaPineda2016SALSAAN} & SALSA & $>$1 groups & Conv. & Standing & A,P & 18 & 1h &  Position, Head, Body Orientation & F-formations \\
         \citealt{Edwards2016FromPT} & CONVERSE & Dyadic & Conv.+ Collab. & Standing & A & 16 & 8h & Body Landmarks, Gaze, Face Expressions & \ding{55} \\
         \citealt{Beyan2016DetectingEL} & Leadership Corpus & Group & Collab. & Seated & A,P,Q & 64 & $\sim$7h & \ding{55} & Leadership \\
         \citealt{Chou2017NNIMETN} & NNIME & Dyadic & Conv. & Seated & A,B,T & 44 & 11h & \ding{55} & Emotions \\
         \bottomrule
    \end{tabular}
         \caption{Datasets that feature audiovisual non-acted social interactions and are publicly available. They are presented grouped by recording setup (third-person, egocentric, and computer-mediated). Abbreviations: Conv. conversational; Collab., collaborative, Comp.; competitive, A, audiovisual; P, psychological; B, biosignals; T, transcriptions; Q, questionnaires; IMU, Inertial measurement unit; ?, value not found; *, robot interaction.}
    \label{tab:datasets1}
\end{table}

\addtocounter{table}{-1}
\begin{table}[!t]
\tiny
    \centering
    \begin{tabular}{>{\raggedright}p{1.7cm}P{1.5cm}P{1.0cm}P{1.0cm}P{1.0cm}P{1.0cm}P{1.0cm}P{1.0cm}P{2.2cm}P{3.0cm}}
         \toprule
         & Dataset & Scenario & Task & Setting & Content & \#Subjects & Size & \multicolumn{2}{c}{Behavioral Annotations} \\
          &  &  &  &  & & & & Low-level & High-level \\
         \midrule
         \multicolumn{10}{c}{Third-person view} \\
         \midrule
         \citealt{Georgakis2017TheCE} & CONFER & Multiple & Comp. & Seated & A & 54 & 2.4h & Face Landmarks, Person Tracking & Conflict Intensity \\
         \citealt{Paggio2017TheDN} & NOMCO & Dyadic & Conv. & Standing & A,T & 12 & $\sim$1h & Head Motion, Face Expressions, Body Landmarks & Emotions, Gestures \\
         \citealt{Bozkurt2017TheJD} & JESTKOD & Dyadic & Conv. & Standing & A,P & 10 & 4.3h & Body Landmarks, Body Motion & Activation, Valence and Dominance \\
         %\citealt{Gu2018AVAAV} & AVA & Multiple & Multiple & Seated & A & ? & 109.2h & Person Tracking & Atomic Action Labels \\
         \citealt{Andriluka2018} & PoseTrack & Multiple & Multiple & Mixed & A & $>$300 & $\sim$45h & Body Landmarks & \ding{55} \\
         \citealt{Marcard2018} & 3DPW & Multiple & Multiple & Mixed & A & ? & 0.5h & 3D Body Landmarks & \ding{55} \\
         \citealt{Mehta2018} & MuPoTS-3D & Dyadic, Triadic & Conv. & Seated & A & 8 & 0.05h & 3D Body Landmarks & \ding{55} \\
         \citealt{Lemaignan2018ThePD} & PInSoRo* & Dyadic & Multiple & Mixed & A & 120 & 45h & Face, Body Landmarks & Task Engagement, Social Engagement, Social Attitude \\
         \citealt{CabreraQuiros2018TheMD} & MatchNMingle & $>$1 groups & Conv. & Seated & A,P & 92 & 2h &  Acceleration, Proximity & Social Actions, Cues and Signals, F-Formations \\
         \citealt{Celiktutan2019MultimodalHI} & MHHRI* & Dyadic, Triadic & Conv. & Seated & A,B & 18 & 4.2h & Wrist Acceleration & Self-Reported Engagement \\
         \citealt{Joo2019PanopticSA} & CMU Panoptic & Triadic, Group & Multiple & Mixed & A,P,T & ? & 5.5h & 3D Body, Face and Hands Landmarks & Speaking Status, Social Formation  \\
         \citealt{Carreira2019ASN} & Kinetics-700 & Multiple & Multiple & Mixed & A & ? & 1805h & \ding{55} & Action Labels \\
         \citealt{Lee2019TalkingWH} & Talking With Hands 16.2M & Dyadic & Conv. & Standing & A & 50 & 50h & Body, Hands Landmarks & \ding{55} \\
         \citealt{Zhao2019HACSHA} & HACS & Multiple & Multiple & Mixed & A & ? & 861h & \ding{55} & Action Labels \\
         \citealt{Monfort2020MomentsIT} & Moments in time & Multiple & Multiple. & Mixed & A & ? & 833h & \ding{55} & Action Labels \\
         \citealt{Chen2020DyadicSA} & DAMI-P2C & Dyadic & Conv. & Seated & A,P,T & 68 & 21.6h & \ding{55} & Parent Perception, Engagement, Affect \\
         \citealt{Maman2020GAMEONAM} & GAME-ON & Triadic & Multiple & Standing & A,Q & 51 & 11.5h & Motion Capture & Cohesion, Leadership, Warmth, Competence, Competitivity, Emotions, Motivation \\
         \citealt{Khan2020VyaktitvAM} & Vyaktitv & Dyadic & Conv. & Seated & A,P,T & 38 & $\sim$6.7h & \ding{55} & Lexical Annotations \\
         \citealt{Schiphorst2020Video2ReportAV} & Video2Report & Dyadic & Conv. (Medical Visit) & Mixed & A & 4 & $\sim$7.1h & Body Landmarks & Action Labels (Medical) \\
         \citealt{Park2020KEmoConAM} & K-EmoCon & Dyadic & Conv. & Seated & A,B & 32 & 2.8h & Accelerometer & \ding{55} \\
         \citealt{Yang2021ADO} & CongreG8 & Triadic, Group & Comp. & Standing & A,P,Q & 38 & $\sim$28h & Motion Capture & \ding{55} \\
         \citealt{MartnMartn2021JRDBAD} & JRDB-Act & Multiple & Multiple & Mixed & A & ? & 1h & Body Point Cloud, Person Detection and Tracking & Atomic Action Labels, Social Formations \\
         \citealt{Doyran2021MUMBAIMM} & MUMBAI & Dyadic & Collab +Comp. & Seated & A,P,Q & 58 & 46h & Body, Face Landmarks & Emotions \\
         \citealt{Palmero2022} & UDIVA v0.5 & Dyadic & Conv.+ Collab. +Comp. & Seated & A,Q,T & 134 & 80h & Face, Body, Hands Landmarks, Gaze & \ding{55} \\
         \bottomrule
    \end{tabular}
         \caption{(Continuation) Datasets that feature audiovisual non-acted social interactions and are publicly available. They are presented grouped by recording setup (third-person, egocentric, and computer-mediated). Abbreviations: Conv. conversational; Collab., collaborative, Comp.; competitive, A, audiovisual; P, psychological; B, biosignals; T, transcriptions; Q, questionnaires; IMU, Inertial measurement unit; ?, value not found; *, robot interaction.}%Continuation of \autoref{tab:datasets1}.}
    \label{tab:datasets2}
\end{table}

\addtocounter{table}{-1}
\begin{table}[!t]
\tiny
    \centering
    \begin{tabular}{>{\raggedright}p{1.7cm}P{1.5cm}P{1.0cm}P{1.0cm}P{1.0cm}P{1.0cm}P{1.0cm}P{1.0cm}P{2.2cm}P{3.0cm}}
         \toprule
         & Dataset & Scenario & Task & Setting & Content & \#Subjects & Size & \multicolumn{2}{c}{Behavioral Annotations} \\
          &  &  &  &  & & & & Low-level & High-level \\
         \midrule
         \multicolumn{10}{c}{First-person view (Egocentric)} \\
         \midrule
         \citealt{Bambach2015LendingAH} & EgoHands & Dyadic & Collab +Comp. & Seated & A & 4 & 1.2h & Hands Segmentation & \ding{55} \\
         \citealt{Yonetani2016RecognizingMA} & PEV & Dyadic & Conv. & Seated & A & 6 & $\sim$0.5h & \ding{55} & Actions and Reactions~Labels \\
         \citealt{Silva2018} & DoMSEV & Multiple & Multiple & Mixed & A & ? & 80h & IMU, GPS & Actions Label \\
         \citealt{Abebe2018AFV} & FPV-O & Multiple & Conv. (Office) & Mixed & A & 12 & 3h & \ding{55} & Actions Label \\
         \citealt{Grauman2021Ego4DAT} & Ego4D & Multiple & Multiple (Daily Life) & Mixed & A & 923 & 3670h &  Gaze & \ding{55} \\
         \midrule
         \multicolumn{10}{c}{Computer-Mediated (Online Interactions)} \\
         \midrule
         \citealt{McKeown2010TheSC} & SEMAINE & Dyadic & Conv. & Seated & A,T & 20 & $\sim$6.5h & \ding{55} & Emotions, Epistemic States, Interaction Actions, Engagement \\
         \citealt{Ringeval2013IntroducingTR} & RECOLA & Dyadic & Collab. & Seated & A,B,P & 46 & 3.8h & \ding{55} & Valence, Arousal, Agreement, Dominance, Performance, Rapport, Engagement, Utterance \\
         \citealt{Cafaro2017TheND} & NoXi & Dyadic & Conv. & Standing & A,T & 87 & 25h &  Body and Face Landmarks, Smiling, Head and Hands Gestures & Arousal, Engagement, Turn Talking \\
         \citealt{Feng2017} & Learn2Smile & Dyadic & Conv. & Seated & A & 500 & $\sim$30h & Face Landmarks & \ding{55} \\
         \citealt{kossaifi2019sewa} & SEWA DB & Dyadic & Conv. & Seated & A,T & 398 & 44h & Face Landmarks and Action Units, Head and Hands Gestures & Valence, Arousal, Liking/Disliking, Agreement, Mimicry, Backchannel, Laughs \\
         \bottomrule
    \end{tabular}
         \caption{(Continuation) Datasets that feature audiovisual non-acted social interactions and are publicly available. They are presented grouped by recording setup (third-person, egocentric, and computer-mediated). Abbreviations: Conv. conversational; Collab., collaborative, Comp.; competitive, A, audiovisual; P, psychological; B, biosignals; T, transcriptions; Q, questionnaires; IMU, Inertial measurement unit; ?, value not found; *, robot interaction.}%Continuation of \autoref{tab:datasets1} and \autoref{tab:datasets2}.}
    \label{tab:datasets3}
\end{table}
\addtolength{\tabcolsep}{5pt} 
\renewcommand{\arraystretch}{1.5}

%We encourage the reader to refer to~\autoref{tab:datasets} for more specific details about the datasets and comparisons among them.

%\color{blue}
We summarize and compare all datasets reviewed in Table~\ref{tab:datasets1}. 
They appear classified into first-person (egocentric), third-person (mid-distance camera), and computer-mediated (e.g., video-conferences) recording setups due to their significant differences regarding their possible applications. 
Datasets recorded from very distant third-person views, or egocentric views where the camera is carried by a non-interacting agent were discarded due to their poor social behavior content. 
Usually, third-view datasets consist of structured interactions where participants need to follow basic directives which favor spontaneous and fluent interactions. Despite the fact that conversations are the most common interaction structure, there are datasets which aim at fostering specific social signals like leadership, competitiveness, empathy, or affect, and therefore engage the participants in competitive/cooperative scenarios \citep{Hung2010TheIW, SanchezCortes2012ANB, Rehg2013DecodingCS, Ringeval2013IntroducingTR, Vella2013OverlapsIM, Bambach2015LendingAH, Salter2015TheTG, Edwards2016FromPT, Beyan2016DetectingEL, Georgakis2017TheCE, Yang2021ADO, Doyran2021MUMBAIMM, Palmero2022}. Other datasets, instead, record in-the-wild interactions during the so-called cocktail parties~\citep{AlamedaPineda2016SALSAAN, CabreraQuiros2018TheMD} and represent very interesting benchmarks to study group dynamics. 
%Few datasets focus on very particular environments such as at the office~\citep{Abebe2018AFV} or at the doctor's office~\citep{Schiphorst2020Video2ReportAV}. 
Thanks to the camera portability during the collection, egocentric datasets can record social behavior in less constrained environments. Very recently, \cite{Grauman2021Ego4DAT} released more than 3000 hours of in-the-wild egocentric recordings of human actions, which also include social interactions.
Finally, the computer-mediated recording setup elicits a very particular behavior due to the idiosyncrasies of the communication channel~\citep{McKeown2010TheSC, Ringeval2013IntroducingTR, Cafaro2017TheND, Feng2017, kossaifi2019sewa}. For example, the latency or the limited field of view might affect the way social cues and signals are transmitted and observed.

With regards to the setting, participants might interact while standing or while seated. Some datasets may include videos with both configurations (e.g., several independent interactive groups).
The most frequent scenario consists in dyadic interactions due to their special interest for human-robot interaction and human behavior understanding, and their lower behavioral complexity when compared to bigger social groups. Triadic~\citep{SanchezCortes2012ANB, Mehta2018, Celiktutan2019MultimodalHI, Joo2019PanopticSA, Yang2021ADO} and bigger social gatherings~\citep{McCowan2005TheAM, Hung2010TheIW, Beyan2016DetectingEL, Joo2019PanopticSA, Yang2021ADO}, which are referred to as \textit{groups}, are less commonly showcased scenarios. Datasets featuring several simultaneous groups of interactions, as long as they showed \textit{focused interactions}, are also included \citep{AlamedaPineda2016SALSAAN, CabreraQuiros2018TheMD}.

Regarding the content released, although originally available, some of them did not release the audio of the videos showcased due to privacy concerns. This is especially frequent for egocentric videos, as the unconstrained recording of the interactions obstructs the collection of consent forms. Other common content typologies consist of psychological data (e.g., personality questionnaires), biosignals monitoring (e.g., heart rate, electrocardiogram, electroencephalograms) and transcriptions. The latter is considerably less frequent due to its tedious manual annotation process~\citep{McCowan2005TheAM, DouglasCowie2007TheHD, McKeown2010TheSC, Lcking2012DatabasedAO, Vella2013OverlapsIM, Vandeventer20154DCC, naim2015automated, Chou2017NNIMETN, Paggio2017TheDN, Cafaro2017TheND, Joo2019PanopticSA, kossaifi2019sewa, Chen2020DyadicSA, Khan2020VyaktitvAM, Palmero2022}. The most frequent low-level annotations that the datasets provide are the participants' body poses and facial expressions~\citep{DouglasCowie2007TheHD, Rehg2013DecodingCS, Bilakhia2015TheMM, Vandeventer20154DCC, naim2015automated, Edwards2016FromPT, Cafaro2017TheND, Feng2017, Georgakis2017TheCE, Paggio2017TheDN, Bozkurt2017TheJD, Andriluka2018, Marcard2018, Mehta2018, Lemaignan2018ThePD, Joo2019PanopticSA, kossaifi2019sewa, Schiphorst2020Video2ReportAV, Doyran2021MUMBAIMM, Palmero2022}. Given their annotation complexity, they are usually automatically retrieved with tools like OpenPose~\citep{cao2019openpose}, and manually fixed or discarded. Others use more complex retrieval systems like motion capture, or mocap~\citep{Edlund2010SpontalAS, Maman2020GAMEONAM, Yang2021ADO}. However, the characteristics of the mocap recording setup and the special suit that participants wear could unintentionally bias the elicited behaviors during the interaction. Finally, the annotation of high-level social signals is often led by the needs of the study for which the dataset was collected. Indeed, some of the datasets have been complementary annotated and added in posterior studies. As a result, most common high-level labels consist of elicited emotions~\citep{McCowan2005TheAM, DouglasCowie2007TheHD, vanSon2008TheIC, McKeown2010TheSC, naim2015automated, Vandeventer20154DCC, Chou2017NNIMETN, Paggio2017TheDN, Maman2020GAMEONAM, Doyran2021MUMBAIMM}, action labels~\citep{Soomro2012UCF101AD, Yonetani2016RecognizingMA, Silva2018, Abebe2018AFV, Carreira2019ASN, Zhao2019HACSHA, Schiphorst2020Video2ReportAV, Monfort2020MomentsIT, MartnMartn2021JRDBAD}, and social cues/signals~
\citep{Hung2010TheIW, SanchezCortes2012ANB, Ringeval2013IntroducingTR, Vandeventer20154DCC, Shukla2016MuDERIMD, Bozkurt2017TheJD, Cafaro2017TheND, Feng2017, Lemaignan2018ThePD, CabreraQuiros2018TheMD, Celiktutan2019MultimodalHI, Chen2020DyadicSA, Maman2020GAMEONAM}.

%\color{black}

%\color{blue} Orden Cronológico/Resume: 
%Más o menos tienes el resumen en la primera columna del excel pero básicamente se clasificaria de la siguiente forma:
%Videos extraidos de Youtube, Escenarios controlados Pie/Sentado, Escenarios con interacción de Robot,
%Actividades cuotidianas/oficina, datasets con cierta finalidad para un objetivo concreto de investigación tipo emociones y algunos datasets utilizan el renderizado de la persona en 3D.

%\begin{itemize}
%\item Todos los datasets son práticamente contenido audiovisual, algunos quitan el audio(marcado con alpha significa que solo hay video).
%\item Las anotaciones al principio eran todo manualmente, pero se han ido utilizando tecnicas automaticas para extraccion de features, o herramientas como Amazon Mechanical Turk.
%\item Las transcripciones realmente se han empezado a hacer en los ultimos años
%\item Se han ido añadiendo dispositivos que añadan multimodalidad como biosignals, GPS...en formato raw data.
%\item Cada vez se ha tenido más en cuenta el multi-view(Pocos datasets tienen un contexto de la escena o persona en 3D)
%\item Muchos datasets tienen cuestionarios de autoevaluación psicológicos o sensor psicologico de expresiones para personalidad.
%\item Realmente hay poco uso de landmarks en los datasets(algunos los extraen para experimentos dentro del paper pero no son incluidos en el dataset), se ha priorizado Action Labels(clasificacion de las acciones dentro del segmento)
%\end{itemize}
%\color{black}

%% file: sections/4_metrics.tex
\section{Metrics}
\label{sec:metrics}

\begin{figure}[!t]
    \centering
    \includegraphics[width=0.8\textwidth]{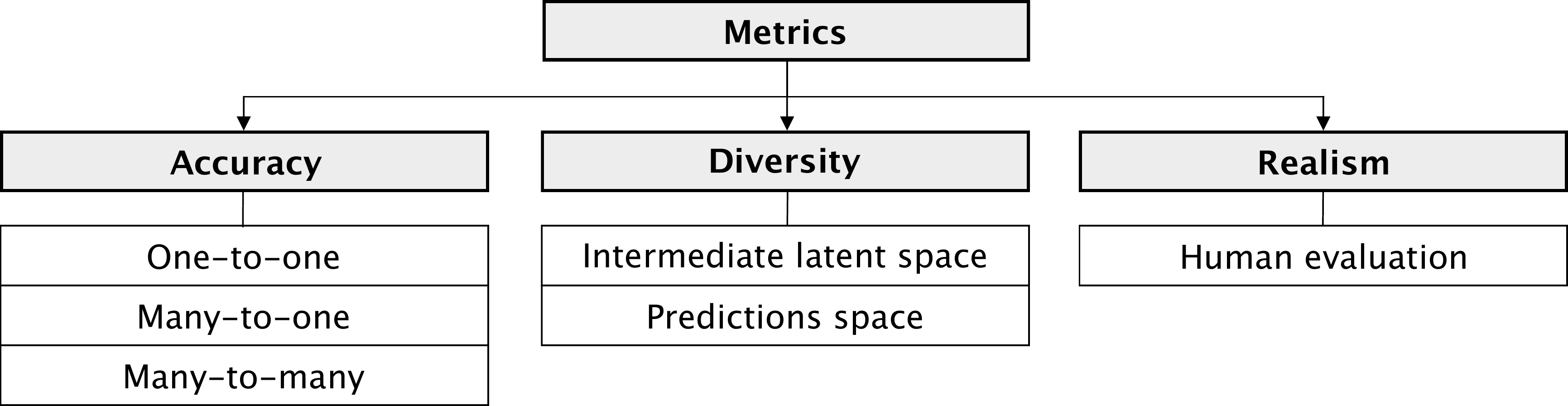}
    \caption{Classification of metrics commonly used in non-verbal social behavior forecasting.}
    \label{fig:metrics_taxonomy}
\end{figure}

%The appropriate evaluation of forecasting methods has always generated discussion. 
The metrics used in behavior forecasting can be clustered in three well-defined branches by the property of the behavior that they assess, see \autoref{fig:metrics_taxonomy}. This classification will be used in this section to present and group the metrics.

On one side, works that predict high-level behavioral representations such as social cues or signals often need to compare single or multiple discrete predictions to a single observed ground truth (\textit{one-to-one}). For univariate classification, the Area Under the Curve (AUC) of the Receiver Operating Curve (ROC) is a common choice, as it very well describes the overall performance of the model. The classic accuracy, precision, recall, and F1-score are valid alternatives for both single and multi-class classification. On the other side, low-level behavioral representations are often continuous and therefore conceived as regression tasks. For those, the L1 and L2 distances have traditionally been the golden standard. Variants of L2-based metrics have been used depending on the field of application. For example, in human motion forecasting with joints, the Mean-Per-Joint-Position Error (MPJPE), the Percentage of Correct Keypoints (PCK), or the cosine similarity are very frequent options. In this context, \cite{Adeli2021} raised the common problem of missing data (e.g., occluded or out-of-view joints). To address this, they proposed metrics that evaluated the models performance under several visibility scenarios: the Visibility-Ignored Metric (VIM), the Visibility-Aware Metric (VAM) and the Visibility Score Metric (VSM). Differently, in behavior forecasting with raw image or audio, quality metrics like the Mean-Squared Error (MSE), the Structural Similarity Index Measure (SSIM)~\citep{wang2004image}, or the Peak Signal to Noise Ratio (PSNR) are better suited. Indeed, this task shares many similarities with video prediction, so any metric from that field can be leveraged for ours~\citep{oprea2020review}.
%Init discussion on whether they are actually appropriated for behavior forecasting}

%F1, MSE, MAE, APE, RMSE
%ADE: average displacement error
%FDE: final displacement error
%MAE + RMSE for top-5/6 AU
%MSE (L2), cosine sim
%L2, Shifted L2, Variation (STD of each dimension of the prediction)
%Evaluate the differences between distributions in attributes extracted from generated faces

A common argument used against accuracy-based metrics is supported on the idea that there are multiple valid and equally plausible futures. For example, using a distance-based metric such as mean-squared error in hands gestures forecasting could end up penalizing models which forecast a gesture over another which could be also suitable to that situation. Similarly, a method could forecast the correct high-frequency gesture but under a small delay and yield a very low accuracy score. To account for such future stochasticity, works embracing this paradigm try to predict all possible future sequences (Section~\ref{sec:future}). In order to make sure that a representative spectrum of all future possibilities is predicted, stochastic approaches need to measure both the  \textit{accuracy} and \textit{diversity} of their predictions~\citep{yuan2020dlow}. Most of these works predict low-level representations that are continuous in the future (e.g., trajectories, poses). As a result, stochastic metrics herein presented may be biased towards such representations. However, most of them are directly applicable to high-level representations forecasting.

The accuracy of methods under the stochastic assumption can be quantified under two different paradigms. The first consists in assessing that at least one of the predicted futures is accurate and matches the ground truth (\textit{many-to-one}). To do so, first, the predicted sample most similar to the ground truth is selected.
%$\frac{1}{T} \min_{x\in X} \lvert \hat{x} - x\rvert$. 
Then, its accuracy can be simply computed with any of the deterministic metrics previously presented (e.g., Average Displacement Error, ADE, ~\citealt{yuan2020dlow}). 
The second, instead of assuming that at least one predicted sample is certain, consists in generating a hypothetical set of multiple ground truths (\textit{many-to-many}). They usually do so by grouping similar past sequences. The future sequences from those grouped observations are considered their \textit{multimodal} ground truth. In this scenario, multimodal adaptations of the deterministic metrics are usually leveraged. Examples include the Multimodal ADE (MMADE), and the Multimodal FDE (MMFDE)~\citep{mao2021generating}. Nonetheless, any one-to-one metric can be computed across all possible futures and then averaged to get a multimodal score. Regarding the quantification of the diversity across the multiple predicted futures, the analysis is usually restricted to either the \textit{predictions space} or an \textit{intermediate latent space}.
For direct comparison in the predictions space, some works use the Average Pairwise Distance (APD), which is calculated as the average L2 distance between all pairs of predictions~\citep{yuan2020dlow, mao2021generating, aliakbarian2021contextually}. The higher this metric is, the higher the variability among predictions.
%$
%\frac{1}{K(K-1)}\sum^K_{j\neq i} \lvert x_i - x_j \rvert
%$.
Similarly, the Average Self Distance (ASD) and the Final Self Distance (FSD) were proposed~\citep{yuan2019diverse}. They both measure, for each future sample from the predictions set $Y$, its minimum distance to the closest future from $Y$ (in terms of L2 distance, for example). All timesteps are averaged for the ASD, and only the last timestep is considered for FSD. These metrics penalize methods sampling repeated futures.
Regarding the diversity-based metrics computed in the latent space, the usual choices are the popular Frechet Inception Distance (FID) and the Inception Score (ID), which are distribution-based metrics that measure the generation fidelity~\citep{yang2019diversity, aliakbarian2020stochastic, cai2021unified}. Additionally, \cite{cai2021unified} presented a \textit{Diversity} score that estimates the feature-based standard deviation of the multiple generated outputs from the same past observation.

Finally, the realism, naturalness, smoothness and plausibility of behavior forecasting is often visually assessed by human raters. To do so, some works prepare questionnaires asking questions like \textit{``Is the purpose of the interaction successfully achieved?"}, \textit{``Is the synthesized agent behaving naturally?"}, \textit{``Does the synthesized agent look like human rather than a robot?"}~\citep{Shu2016}.
The answers are usually scales of discrete values (e.g., between 0 and 5). During the evaluation process, those questions are presented while showing a behavior sampled either from the ground truth or from the predictive model~\citep{kucherenko2021large}. This prevents the introduction of biases in the humans' ratings. The posterior usage of statistical hypothesis tests helps to conclude whether the predicted behavior is humanlike~\citep{Feng2017, Chu2018, Woo2021}.
Unfortunately, the amount of unique questions used in the literature is very large, with very few of them being repeated across studies~\citep{fitrianie2019we}. There have been few attempts to create a unified measurement instrument that helps reduce such heterogeneity~\citep{fitrianie202019, fitrianie2021questionnaire}. Alternatively, other works use pairwise evaluations. In those, the human rater selects the most human-like behavior among a pair of samples. A recent study observed the superior inter-rater reliability of pairwise evaluations, which favored it over questionnaires for small-scale studies~\citep{wolfert2021rate}.
Overall, most subjective evaluations lack a systematic approach that ensures high methodological quality. As a result, the extraction of systematic conclusions is often difficult~\citep{wolfert2021review}. 
Additionally, one must be aware though of the possible biases induced by this type of assessments. First, the sampling of the subjects that participate in the qualitative evaluation may include biases towards certain subgroups. For example, the participants selection of a qualitative analysis performed by a graduate student may contain inherent biases towards people from academia. On the other side, the bias could be originated in the choice of the samples to evaluate, the way they are displayed, or even the order in which they are presented. Although most of them are not completely avoidable, they must be taken into consideration and minimized if possible.

%Inside the qualitative part, we need to explain how HRI community assess the results on backchanneling/social signs generation, and how the virtual ones assess it (realism, etc.)

%Regarding the evaluation of the deterministic scenario, \cite{Chu2018} refers to it as the \textit{The Mind-Reading Test}, and acknowledges the limitations of evaluating a sample from the set of plausible future behaviors. 

%% file: sections/5_discussion.tex
\section{Discussion}
\label{sec:discussion}

%\textbf{Taxonomy. }
The forecasting of low-level representations like landmarks or facial action units has been recently tackled with deep learning methods such as recurrent neural networks, graph neural networks, and transformers. The usage of such deep and data-hungry models has been encouraged by the recent availability of large-scale multi-view datasets, often annotated in a semi-automatic way~\citep{Joo2019PanopticSA, Palmero2022}. The increasing accuracy of monocular and multi-view automated methods for face, pose, and hands estimation has contributed in reducing the annotation effort. Still, the largest available datasets that provide thousands of hours of audiovisual material and feature the widest spectrum of behaviors do not provide such annotations~\citep{Carreira2019ASN, Zhao2019HACSHA, Monfort2020MomentsIT, Grauman2021Ego4DAT}. In contrast, the automated methods for high-level representations recognition such as feedback responses or atomic action labels are not accurate enough to significantly help in their annotation procedures. Consequently, such annotations are scarce, and are only available for small datasets, as shown in our survey. Accordingly, recent works have opted for classic methods such as SVM, AdaBoost, and simple recurrent neural networks, which have traditionally worked fairly well with small datasets. We expect future work on high-level behavior forecasting to also explore semi- and weakly-supervised approaches~\citep{Jain2021}.

Latest works that focus on forecasting low-level representations have proposed methods that successfully exploit interpersonal dynamics in very specific scenarios (e.g., dancing) by using cross-interaction attention~\citep{Katircioglu2021, Guo2021}. In other tasks where these dynamics may not be so strong and frequent (e.g., conversational), or simply not sufficiently captured by the visual cue chosen (e.g., landmarks), the adequate inclusion of such features still has further room for improvement~\citep{Barquero2022}. In fact, the influence of the scenario and the input representation on performance is also observed in works that focus on forecasting high-level representations. For instance, using head and body pose as input and output, each represented by a selected 3D landmark coordinate and a normal,~\citet{Raman2021} observed that the addition of features from the other interlocutors harmed performance for the three predicted categories (including speaking status) when applied to an in-the-wild mingling scenario. In contrast, in a structured triadic scenario, head and body location were better predicted when adding such information (MSE of up to 15.84 cm vs. 18.20 cm), whereas orientation was better predicted without. When using input features other than landmarks, most works tend to benefit from the addition of the partner's features, even in conversational scenarios. For instance,~\citet{Ishii2021} showed consistently superior performance when using speaker and listener features compared to using just one of them for backchannel (F1 score of 85.2\% vs. up to 74.2\%) and turn-changing (F1 score of 61.7\% vs. up to 59.2\%) prediction in a seated dyadic scenario.

%This is consistent with other works that focus on forecasting high-level 
%The latter is further supported by findings from the social cue forecasting literature.

With respect to the methodological trends of low-level representation forecasting, we foresee a bright future for methods that use representations in the frequency space. Very recent works have reported promising results with such architectures, especially by helping to alleviate the very limiting freezing motion commonly observed in deterministic approaches. Related to the future perspective, we also expect future works to explore the stochastic point of view. Although many works from the single human behavior forecasting field have already found benefits from assuming the future stochasticity and thus predicting several futures, their translation to social settings remains unexplored. We think that the implementation of socially aware adversarial losses, like the dual-stream one presented by \citet{Airale2021} for behavioral action forecasting, could help to build systems capable of generating diverse, plausible, and socially coherent behavior. 
There are other research lines that have provided many benefits in other fields that also remain uninvestigated. First, the exploration of multimodal approaches has been timidly addressed in the past only to generate immediate next motions~\citep{Hua2019, Ahuja2019}, or in very preliminary and naive ways~\citep{Barquero2022}. Future works should also test self-supervised learning techniques, which have shown their power in conceptually close applications like video prediction
~\citep{oprea2020review}, and could be similarly applied to this field. Furthermore, models that update the learnt behavioral model according to each person's individual behavioral patterns via meta-learning are also very promising~\citep{moon2021fast}.

A popular question when reading works that tackle forecasting problems like ours is whether the results could be transferred to real applications. We all like to see state-of-the-art methods on top of benchmarks and directly choose them for our target application. Instead, we should stop right before making a choice and ask ourselves: \textit{do those numeric accuracy values suit real life scenarios?} Obviously, the answer is not straightforward, and greatly depends to a large degree on the amount of error assumable in each application. 
For example, terminating an interaction during a life-threatening human-robot assistance (e.g., assisted surgery, rescues) does not have room for errors, while doing it during a shopping assistance does. In order to assess this adequacy of the model performance to the real target task, works on social cues/signals forecasting (e.g., backchannel opportunity prediction, interaction breakdown) often perform objective and subjective studies by means of robotic or virtual agents~\citep{Huang2020, Murray2021}. For example, these tests have been used to prove the great capacities of models that generate backchannel responses when it comes to successfully keeping the user engaged during human-robot interactions~\citep{Murray2021}. 
For low-level representations though, there is still a lack of extensive studies that assess the transferability of the results to the target scenario. This is mainly due to the extra constraints posed by these low-level representations. %For example, the actuators of a robot may need some extra milliseconds to perform the actions proposed by the behavior forecasting model. 
In fact, some works highlight the current possibilities and limitations of behavior forecasting with such representations. For example, future behavior in competitive interactions like fencing strongly depends on the player's decisions in answer to the competitor's. Within this context, \cite{Honda2020} showed inferior performance for the rapid and highly stochastic motion of the dominant arm (PCKs of 71.8\% and 66.4\% for dominant hand and elbow, respectively) than for the other parts of the body (average PCK of 77.1\%). This has been consistent in the literature for other less interactive scenarios like face-to-face conversations. \cite{Barquero2022} showed superior accuracy for behavior forecasting for face and upper body torso (errors of 12.70 and 5.75 pixels in average for a prediction of 2 seconds) than for hands (25.15 pixels). This represents an important bottleneck for hands forecasting, where research is almost nonexistent despite of their importance in human communication. The authors also showed superior performance (error of 5.34 pixels) than a naive but strong baseline (6.00 pixels) for the short term ($<$400ms). This opens new possibilities for providing human-robot interactive agents with fairly accurate anticipation capabilities. For example, the proper activation of the actuators of a robot may benefit from any extra milliseconds of anticipation. 
In general though, we think that landmarks-based behavior forecasting is still immature, and will strongly benefit from further research efforts. Another concerning issue related to this topic lies on the typology of the data leveraged for forecasting. Only models that make their predictions solely leveraging automatically retrieved data can be successfully applied to real life scenarios. Actually, and similarly to the low-level scenario, we expect the forecasting of high-level behavioral representations to greatly benefit from the development of new accurate methods to automatically retrieve social cues/signals from raw image/audio data. 

Finally, we want to raise awareness of what we consider one of the main bottlenecks of behavior forecasting: the evaluation metrics. An evaluation metric must always illustrate how well a method does for the target task. While this sentence may seem trivial when thinking of classic classification or regression tasks, it is an important source of controversy in the behavior forecasting field. For instance, the distance between the generated and the ground truth futures does not describe the coherence of the pose in all future steps, neither the realism of the movements. In fact, it does not even guarantee that a method with low error performs poorly, as the predictions may simply not match the ground truth, which is a sample of the multiple and equally plausible set of futures. Although one would conclude that a proper evaluation always contains a qualitative analysis, multiple behavioral dimensions may escape from human raters and therefore bias it. For instance, it is not trivial to build a qualitative analysis that also assesses the coherence of the predicted behavior with respect to the behavioral patterns specific to the subject, the context, or even the events from the mid- to long-term past. We hope that the recent appearance of behavior forecasting benchmarks and specific datasets will encourage the community to find better-suited metrics and evaluation protocols that will boost the research progress in this field.

%% file: sections/6_conclusion.tex
\section{Ethics} 
\label{sec:ethics}

% TO BE DISCUSSED
% 1) Applications for good: pedagogical agents (e-learning), assistive robots (e-health), collaborative tasks (surgery).
% 2) Ethical concerns: non-consensual behavior forecasting, unfair algorithms (due to biases), manipulation/persuasion that these systems may induce on their users. E.g., engagement increase
% 3) Adversarial attacks: users may exploit the system's vulnerabilities in their benefit.
% 4) Understandable AI: transparency. Users engaging with behavior forecasting agents need to be aware of such practices.

 We have discussed many applications for good where non-verbal social behavior forecasting might be valuable. Personalized pedagogical agents~\citep{davis2018} that maximize learner's attention and learning, empathetic assistive robots for hospital patients or dependent people~\citep{andrist2015, esterwood2021}, and collaborative robots for industrial processes or even surgeries~\citep{sexton2018anticipation} are few examples. However, each new technology comes with its own pitfalls and limitations. 
In fact, these algorithms may unintentionally hold important biases that lead to unfairness in the task being performed. For example, the implementation of behavior forecasting algorithms in security borders or migration controls might lead to undesired outcomes~\citep{mckendrick2019} interfering with human rights~\citep{ahkmetova2021}. 
Furthermore, the interacting user should always be aware of the presence of such forecasting systems, the possible manipulations or persuasion techniques attached, and their ultimate goal. Unfortunately, providing the user with these descriptions is not always easy because most of the times such systems are neither transparent nor explainable. Therefore, the incorporation of specific techniques to promote such interpretability is of utmost importance in order to build trust with the user. 
On the other side, it is also important to consider the potential vulnerabilities that such systems may have and how users might exploit them driven by unethical purposes. This is especially important for assistive or collaborative robots, which often involve very sensitive scenarios. 
Finally, although data protection regulations vary across countries~\citep{guzzo2015}, data privacy and data protection must ensure informational self-determination and consensual use of the information that can be extracted with the methods presented herein. In this sense, frameworks such as the EU General Data Protection Regulation (GDPR\footnote{\url{https://gdpr.eu/}.}) provide excellent safeguards for establishing ethical borders that should not be crossed.

\section{Conclusion}

In this survey, we provided an overview of the recent approaches proposed for non-verbal social behavior forecasting. We formulated a taxonomy that comprises and unifies recent (since 2017) attempts of forecasting low- or high-level representations of non-verbal social behavior up to date. By means of this taxonomy, we identified and described the main challenges of the problem, and analyzed how the recent literature has addressed them from both the sociological and the computer vision perspectives. We also presented all audiovisual datasets related to social behavior publicly released up to date in a summarized, structured, and friendly way. Finally, we described the most commonly used metrics, and the controversy that they often raise. We hope this survey can help bring the human motion prediction and the social signal forecasting worlds together in order to jointly tackle the main challenges of this field.